%% file: main.tex
\documentclass[10pt,journal,compsoc]{IEEEtran}
\usepackage{cite}
\ifCLASSINFOpdf
\else
\fi

\usepackage{pifont}
\usepackage{subfiles}
\usepackage{graphicx}
\usepackage[colorlinks]{hyperref} 
\usepackage{subfigure}
\usepackage[labelfont=bf]{caption}
\usepackage{amsfonts}
\usepackage{mathtools}
\usepackage{booktabs}
\usepackage{multirow}
\usepackage{cite}
\usepackage{comment}
\newcommand{\Paragraph}[1]{\vspace{1mm} \noindent \textbf{#1.} \hspace{0mm}}

\usepackage{setspace}
\usepackage{algorithm}
\usepackage{algorithmic} 
\usepackage[ruled,vlined,algo2e]{algorithm2e}
\begin{document}

\title{Learning Implicit Functions for Dense 3D Shape Correspondence of Generic Objects}

\author{Feng~Liu,~\IEEEmembership{Member,~IEEE,}
        and~Xiaoming~Liu,~\IEEEmembership{Fellow,~IEEE}
\IEEEcompsocitemizethanks{\IEEEcompsocthanksitem F.  Liu  and  X.  Liu  are  with  the  Department  of  Computer  Science  and Engineering, Michigan State University, East Lansing, MI 48824, U.S.A.\protect\\
E-mail: \{liufeng6,liuxm\}@msu.edu}}

\markboth{Journal of \LaTeX\ Class Files,~Vol.~14, No.~8, August~2015}%
{Shell \MakeLowercase{\textit{et al.}}: Bare Demo of IEEEtran.cls for Computer Society Journals}

%% ---------------------------------------- %% 
\IEEEtitleabstractindextext{%
\begin{abstract}
The objective of this paper is to learn dense 3D shape correspondence for topology-varying generic objects in an unsupervised manner. Conventional implicit functions estimate the occupancy of a 3D point given a shape latent code. Instead, our novel implicit function produces a probabilistic embedding to represent each 3D point in a part embedding space. Assuming the corresponding points are similar in the embedding space, we implement dense correspondence through an inverse function mapping from the part embedding vector to a corresponded 3D point. Both functions are jointly learned with several effective and uncertainty-aware loss functions to realize our assumption, together with the encoder generating the shape latent code. During inference, if a user selects an arbitrary point on the source shape, our algorithm can automatically generate a confidence score indicating whether there is a correspondence on the target shape, as well as the corresponding semantic point if there is one. Such a mechanism inherently benefits man-made objects with different part constitutions. The effectiveness of our approach is demonstrated through unsupervised 3D semantic correspondence and shape segmentation.
\end{abstract}

}

\maketitle
\IEEEdisplaynontitleabstractindextext
\IEEEpeerreviewmaketitle

%%% ---------------------- ------------------- ----------------- %%

\subfile{sec_1_intro.tex}

\subfile{sec_2_prior.tex}
%%
\subfile{sec_3_method.tex}

\subfile{sec_4_exp.tex}

\subfile{sec_5_conclusion.tex}
%%

%%% ---------------------- ------------------- ----------------- %%

\ifCLASSOPTIONcaptionsoff
  \newpage
\fi

{\small
\bibliographystyle{IEEEtran}
\bibliography{refs}
}
\subfile{sec_6_bio.tex}

\end{document}

%% file: sec_1_intro.tex
\IEEEraisesectionheading{\section{Introduction}\label{sec:intro}}

%% -------------------
\IEEEPARstart{F}{inding} dense correspondence between $3$D shapes is a key algorithmic component in problems such as statistical modeling~\cite{blanz2003face,zuffi20173d,bogo2014faust}, cross-shape texture mapping~\cite{kraevoy2003matchmaker}, and space-time $4$D reconstruction~\cite{niemeyer2019occupancy}.
Dense $3$D shape correspondence can be defined as: 
{\it given two $3$D shapes belonging to the same object category, one can match an arbitrary point on one shape to its semantically equivalent point on another shape if such a correspondence exists}. 
For instance, given two chairs, the dense correspondence of the middle point on one chair's arm should be the similar middle point on another chair's arm, despite different shapes of arms; or alternatively, declare the non-existence of correspondence if another chair has no arm.

The dense $3$D correspondence problem is difficult because it involves understanding the shapes at both the local and global levels.
Prior dense correspondence methods~\cite{ovsjanikov2012functional,litany2017deep,groueix20183d,halimi2019unsupervised,roufosse2019unsupervised,lee2019dense,steinke2007learning,liu20193d} have proven to be effective on organic shapes, \emph{e.g.}, human bodies and mammals. 
However, those methods become less suitable for generic topology-varying or man-made objects, \emph{e.g.}, chairs or vehicles~\cite{huang2014functional}.

\begin{figure*}[t]
\centering    
\subfigure[]{\label{fig:teaser_a}\includegraphics[trim=0 2 0 2,clip, height=55mm]{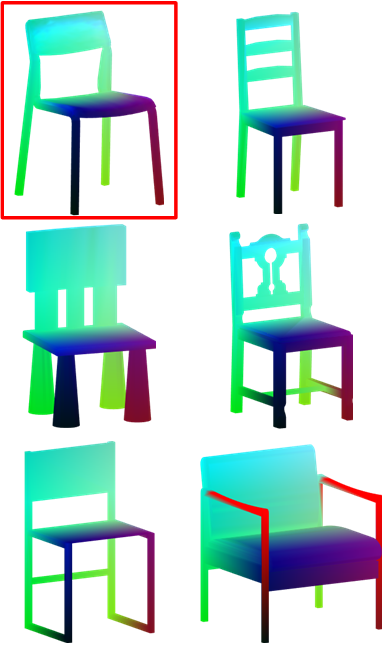}} 
\hspace{5mm}
\subfigure[]{\label{fig:teaser_b}\includegraphics[trim=0 2 0 2,clip,height=55mm]{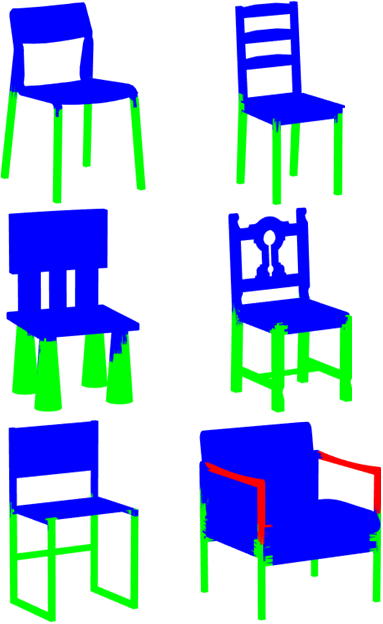}}
\hspace{5mm}
\subfigure[]{\label{fig:teaser_c}\includegraphics[trim=0 9 0 0,clip,height=55mm]{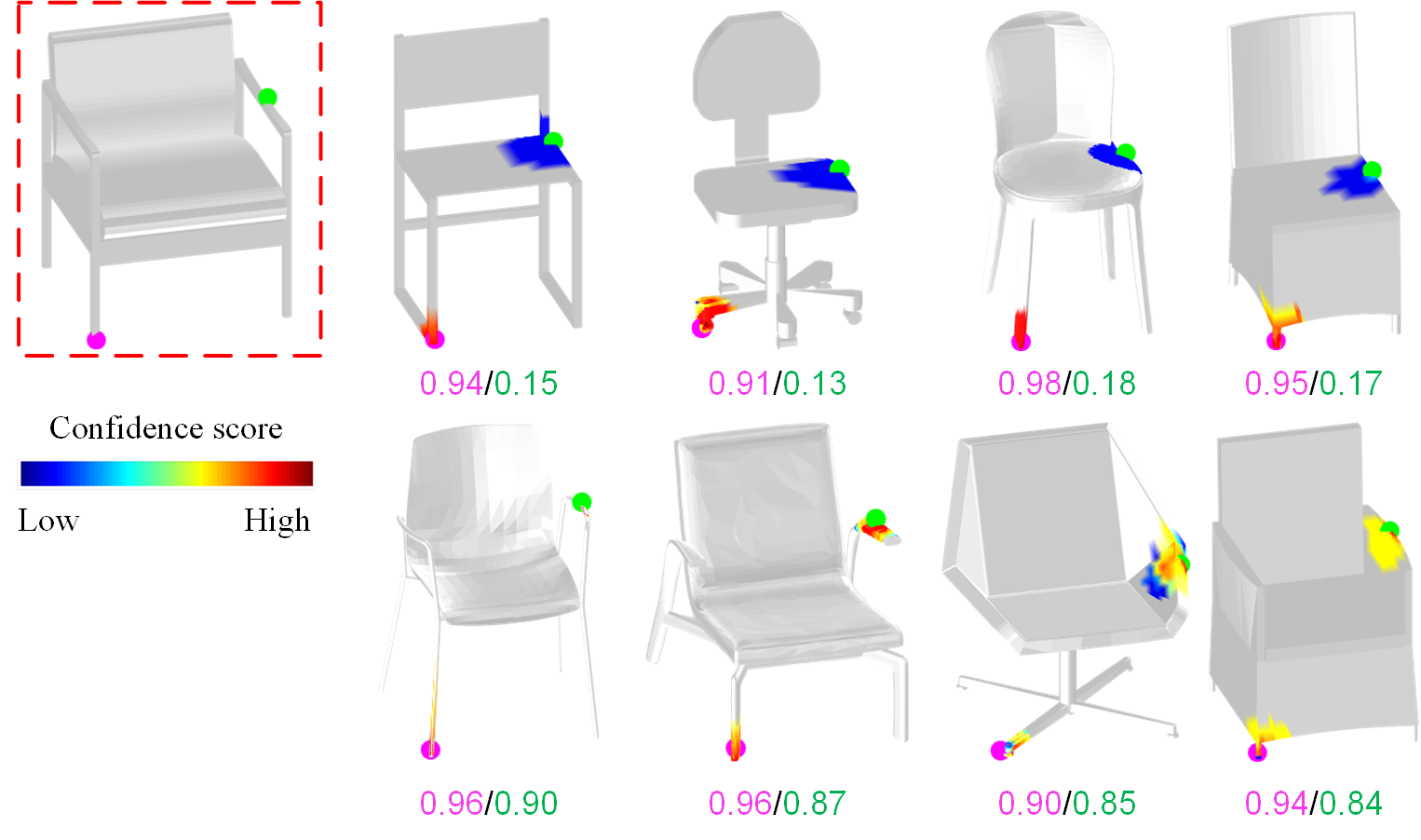}}
\caption{Given a set of $3$D shapes, our category-specific unsupervised method learns pair-wise dense correspondence (a) between any source and target shape (red box), and shape segmentation (b). Give an \emph{arbitrary} point on the source shape (red box), our method predicts its corresponding point on any target shape, and a score measures the correspondence confidence (c).
    For each target, we show the confidence scores of \textcolor{red}{red}/\textcolor{green}{green} points and score maps around corresponded points.
    A score less than a threshold ({\it e.g.}, $0.2$) deems the correspondence as ``non-existing''-- a desirable property for topology-varying shapes with missing parts, {\it e.g.}, chair's arm.}
\label{fig:teaser}
\end{figure*}

%% -------------------
It remains a challenge to build dense $3$D correspondence for a generic object category with large variations in geometry, structure, and even topology. 
First of all, the lack of annotations on dense correspondence often leaves {\it unsupervised learning} the only option. 
Second, most prior works make an inadequate assumption~\cite{van2011survey} that there is a similar topological variability between matched shapes.
Man-made objects such as chairs shown in Fig.~\ref{fig:teaser} are particularly challenging to tackle, since they often differ not only by geometric deformations but also by {\it part constitutions}. 
In these cases, existing correspondence methods for man-made objects either perform fuzzy~\cite{kim2012exploring,solomon2012soft} or part-level~\cite{sidi2011unsupervised,alhashim2015deformation} correspondences or predict a constant number of semantic points~\cite{huang2017learning,chen2020unsupervised}. 
As a result, they cannot determine whether the established correspondence is a ``missing match'' or not. 
As shown in Fig.~\ref{fig:teaser_c}, for instance, we may find non-convincing correspondences in legs between an office chair and a $4$-legged chair, or even no correspondences in arms for some pairs.
Ideally, given a query point on the source shape, a dense correspondence method should be able to determine whether there exists a correspondence on the target shape, and identify the corresponding point if there is. 
This objective lies at the core of this work.

 %\textcolor{red}{red}/\textcolor{green}{green}

%% -------------------
Shape representation is highly relevant to, and can impact, the approach of dense correspondence.
Recently, compared to point cloud~\cite{achlioptas2018learning,qi2017pointnet,qi2017pointnet++} or mesh~\cite{groueix2018atlasnet,meshrcnn,wang2018pixel2mesh}, deep implicit functions have shown to be highly effective as $3$D shape representations~\cite{park2019deepsdf,mescheder2018occupancy,liu2019learning,saito2019pifu,chen2019bae,chen2018learning,atzmon2019controlling}, since it can handle generic shapes of arbitrary topology, which is favorable as a representation for dense correspondence.
Often learned as a multilayer perceptron (MLP), conventional implicit functions input the $3$D shape represented by a latent code $\mathbf{z}$ and a query location $\mathbf{x}$ in the $3$D space, and estimate its occupancy $O=f(\mathbf{x}, \mathbf{z})$. 

In this work, we propose to plant the dense correspondence capability into the implicit function by learning a semantic part embedding. Specifically, we first adopt a branched implicit function~\cite{chen2019bae} to learn a part embedding vector (PEV), $\mathbf{o}=f(\mathbf{x}, \mathbf{z})$, where the max-pooling of PEV $\mathbf{o}$ gives the occupancy $O$.
In this way, each branch is tasked to learn a representation for one universal part of the input shape, and PEV represents the occupancy of the point w.r.t.~all the branches/semantic parts. 
By assuming that PEVs between a pair of corresponding points are similar, we then establish dense correspondence via an inverse function $\mathbf{\hat{x}}=g(\mathbf{o}, \mathbf{z})$ mapping the PEV back to the $3$D space. 
To further satisfy the assumption, we devise an unsupervised learning framework with a joint loss measuring both the occupancy error and shape reconstruction error between $\mathbf{{x}}$ and $\mathbf{\hat{x}}$. 
In addition, a cross-reconstruction loss is proposed to enforce part embedding consistency by mapping between a pair of shapes in the collection. 
Besides, 
we adopt a probabilistic solution to the semantic part embedding learning, where each $3$D point is represented as a Gaussian distribution in the semantic latent space. The mean of the distribution encodes the most likely PEVs while the variance shows the uncertainty along each feature dimension of PEVs.
During inference, the ``likelihood'' between two Gaussian distributions
can then be naturally derived to produce a confidence score for measuring the accuracy of the established point-to-point correspondence. And the learned uncertainty can be interpreted as the model's confidence along each feature dimension of PEVs, which can visualize the distribution of the ``hard'' points of a shape for dense correspondence.

%% -------------------
A preliminary version of this work was published in the $34$th Annual Conference on Neural Information Processing Systems (NeurIPS) $2020$~\cite{liu2020correspondence}. We further extend the work from three aspects: (i) we design a novel \textbf{\emph{deep}} branched implicit function, which gives a more powerful shape representation and improves shape correspondences.
(ii) Instead of representing each point as a deterministic point in the semantic embedding space, we propose to use probabilistic embeddings for semantic part feature learning, which enables our framework to inherently capture the uncertainty of each point correspondence by its probabilistic PEV.
and (iii) we further carry out dense correspondence evaluation and comparison on real scans ($3$D body shapes from Faust dataset~\cite{bogo2014faust}); 

In summary, this paper makes these contributions.                             
\begin{itemize}
 \item We propose a novel paradigm leveraging the  implicit function representation for category-specific \emph{unsupervised and uncertainty-aware} dense $3$D shape correspondence, applicable to generic objects with diverse variations.

 \item We devise several effective loss functions to learn a semantic part embedding, which enables both shape segmentation and dense correspondence. Based on the learned probabilistic part embedding, our method further produces a confidence score indicating whether the predicted correspondence is valid.

 \item Through extensive experiments, we demonstrate the superiority of our method in $3$D shape segmentation, $3$D semantic correspondence, and dense $3$D shape correspondence.  

\end{itemize}

The rest of this paper is organized as follows. Section $2$ briefly reviews related work in the literature. Section $3$ introduces in detail the proposed unsupervised dense $3$D shape correspondence algorithm based on implicit shape representation and its implementations. Section $4$ reports the experimental results. Section $5$ concludes the paper.

%% file: sec_2_prior.tex
\begin{table*}[t]
\renewcommand\arraystretch{1.1}
\newcommand{\greencheck}{\textcolor{green}{\ding{51}}} 
\newcommand{\redX}{\textcolor{red}{\ding{55}}}
\newcommand{\tabincell}[2]{\begin{tabular}{@{}#1@{}}#2\end{tabular}}
\centering
\caption{A comparison of shape correspondence methods designed for generic objects is presented. 'Template' refers to methods that are template-based or require templates. Note that the methods requiring templates and template-based methods might not be  suitable for man-made objects, since they have significant differences in the number and arrangement of their parts.}
\resizebox{0.99\linewidth}{!}{
\begin{tabular}{l |c| |c |c |c | c | c | c }
\toprule
Method & Type & Supervision & \tabincell{c}{Template}  &  \tabincell{c}{Corr.\\ Level}  &  \tabincell{c}{Shape \\ Representation}  & \tabincell{c}{Non-Existence \\ Detection}  & Content, Uncertainty-aware\\
\hline\hline
Slavcheva~\emph{et al.}~\cite{slavcheva2017towards} &  Optimization & unsupervised & \redX & dense & implicit function & \redX  &  bodies, \redX \\
$3$D-CODED~\cite{groueix20183d} & Learning & self-supervised & \greencheck  & dense &  mesh & \redX &   bodies, \redX  \\
LoopReg~\cite{bhatnagar2020loopreg} & Learning & self-supervised & \redX  & dense &  implicit function & \redX &   bodies, \redX  \\
Kim12~\cite{kim2012exploring}  & Optimization & unsupervised & \redX & dense & point & \redX  & man-made objects, \redX  \\
Kim13~\cite{kim2013learning}   & Optimization & unsupervised & \greencheck & dense & point & \redX  & man-made objects, \redX  \\
LMVCNN~\cite{huang2017learning} & Learning & supervised & \redX & dense & point & \redX & man-made objects, \redX    \\
ShapeUnicode~\cite{muralikrishnan2019shape} & Learning & supervised & \redX & sparse & point & \redX & man-made objects, \redX    \\
Chen~\emph{et al.}~\cite{chen2020unsupervised} & Learning & unsupervised & \redX  & sparse &  point & \redX & man-made objects, \redX  \\ 

DIF~\cite{deng2020deformed} & Learning & unsupervised & \greencheck  & dense &  implicit function & \redX & man-made objects, \redX \\ 
Zheng~\emph{et al.}~\cite{zheng2020deep} & Learning & unsupervised & \greencheck  & dense &  implicit function & \redX & man-made objects, bodies, \redX \\ \hline

Proposed  & Learning & unsupervised & \redX  & dense &  implicit function & \greencheck & man-made objects, bodies, \greencheck  \\
\bottomrule
\end{tabular}
}
\label{tab:correspondence_review}
\end{table*}

\section{Related Work}\label{sec:prior}

%% -------------------
\subsection{Dense Shape Correspondence} 
While there are many dense correspondence works for organic shapes~\cite{ovsjanikov2012functional,litany2017deep,groueix20183d,halimi2019unsupervised,roufosse2019unsupervised,lee2019dense,boscaini2016learning,steinke2007learning,shape-my-face-registering-3d-face-scans-by-surface-to-surface-translation}, here we focus on methods designed for man-made objects, including optimization and learning-based methods. 
For the former, most prior works build correspondences only at a {\it part} level~\cite{kalogerakis2010learning,huang2011joint,sidi2011unsupervised,alhashim2015deformation,zhu2017deformation}.
Kim \emph{et al.}~\cite{kim2012exploring} propose a diffusion map to compute point-based ``fuzzy correspondence'' for every shape pair.  
This is only effective for a small collection of shapes with limited shape variations.
~\cite{kim2013learning} and~\cite{huang2015analysis} present a template-based deformation method, which can find point-level correspondences after rigid alignment between the template and target shapes.
However, these methods only predict coarse and discrete correspondence, leaving the structural or topological discrepancies between matched parts or part ensembles unresolved.

%% -------------------
A series of learning-based methods~\cite{yi2017syncspeccnn,huang2017learning,sung2018deep,muralikrishnan2019shape,you2020keypointnet} are proposed to learn local descriptors, and treat correspondence as $3$D semantic landmark estimation.
For example, ShapeUnicode~\cite{muralikrishnan2019shape} learns a unified embedding for $3$D shapes and demonstrates its ability in correspondence among $3$D shapes. 
However, these methods require {\it ground-truth} pairwise correspondences for training. 
Recently, Chen \emph{et al.}~\cite{chen2020unsupervised} present an unsupervised method to estimate $3$D structure points. 
Unfortunately, it estimates a \emph{constant} number of \emph{sparse} structured points. 
As shapes may have diverse part constitutions, it may not be meaningful to establish the correspondence between all of their points.
Groueix~\emph{et~al.}~\cite{groueix2019unsupervised} also learn a parametric transformation between two surfaces by leveraging cycle consistency, and apply it to the segmentation problem. 
However, the deformation-based method always deforms all points on one shape to another, even for the points from a non-matching part.
In contrast, our {\it unsupervised and uncertainty-aware} learning model can perform pairwise {\it dense} correspondence for any two shapes of a man-made object. We summarize the comparison in Tab.~\ref{tab:correspondence_review}.

%% -------------------
\subsection{Implicit Shape Representation} Due to the advantages of being a continuous representation and handling complicated topologies, implicit functions have been adopted for learning representations for $3$D shape generation~\cite{chen2018learning,mescheder2018occupancy,park2019deepsdf,liu2019learning}, encoding texture~\cite{oechsle2019texture,sitzmann2019scene,saito2019pifu}, 3D reconstruction~\cite{fully-understanding-generic-objects-modeling-segmentation-and-reconstruction, voxel-based-3d-detection-and-reconstruction-of-multiple-objects-from-a-single-image, 2d-gans-meet-unsupervised-single-view-3d-reconstruction}, and $4$D reconstruction~\cite{niemeyer2019occupancy}. 
Meanwhile, some extensions have been proposed to learn deep structured~\cite{genova2019learning,genova2019deep} or segmented implicit functions~\cite{chen2019bae}, or separate implicit functions for shape parts~\cite{paschalidou2020learning}. 
Further, some works~\cite{huang2004hierarchical,huang2006shape,slavcheva2017towards,deng2020deformed,zheng2020deep,yenamandra2020i3dmm,bhatnagar2020loopreg} leverage the implicit representation together with a deformation model for shape registration. 
However, these methods rely on the deformation model, which might prevent their usage for topology-varying objects. 
Slavcheva~\emph{et~al.}~\cite{slavcheva2017towards} implicitly obtain correspondence for organic shapes by predicting the evolution of the signed distance field. 
However, as they require a Laplacian operator to be invariant, it is limited to small shape variations. 
Recently, Zheng~\emph{et al.}~\cite{zheng2020deep} present a deep implicit template, a new $3$D shape representation that factors out the implicit template from deep implicit functions. 
Additionally, a spatial warping module deforms the template's implicit function to form specific object instances, which reasons dense correspondences between different shapes. 
Similarly, DIF~\cite{deng2020deformed} introduces a deep implicit template field together with a deformation module to represent $3$D models with correspondences. 
However, these methods assume that the object instances within a category are mostly composed of a few common semantic structures, which inevitably limits their effectiveness for topology-varying objects. 

On the other hand, in order to preserve fine-grained shape details in implicit function learning, Wang~\emph{et al.}~\cite{xu2019disn} propose to combine $3$D query point features with local image features to predict the SDF values of the $3$D points, which is able to generate shape details. 
Meanwhile, instead of encoding the shape in a single latent code $\mathbf{z}$,  Chibane~\emph{et~al.}~\cite{chibane20ifnet} and Peng~\emph{et~al.}~\cite{peng2020convolutional} propose to extract a learnable multi-scale tensor of deep features. Then, instead of classifying point coordinates $\mathbf{x}$ directly, they classify deep features extracted at continuous query points, preserving local details. In this paper, we propose a deep branched implicit function, which can also improve the fidelity of shape representations.

\subsection{Uncertainty in Deep Learning}
Recent years have witnessed a trend to estimate uncertainty in deep neural networks (DNNs)~\cite{blundell2015weight,kendall2017uncertainties,malinin2018predictive,van2020uncertainty}. Specific to deep learning models for the computer vision field, the uncertainties can be classified into two main types: model (or epistemic) uncertainty and data (or aleatoric) uncertainty. Model uncertainty accounts for uncertainty in the model parameters and can be remedied with sufficient training data~\cite{neal2012bayesian,gal2016dropout,kendall2015bayesian}. Data uncertainty captures the noise inherent in the training data, which cannot be reduced even with enough data~\cite{kendall2017uncertainties}.
Recently, uncertainty learning has been widely applied to various tasks, 
such as semantic segmentation~\cite{kendall2015bayesian,huang2018efficient}, depth estimation~\cite{poggi2020uncertainty,ke2021deep}, depth completion~\cite{depth-completion-with-twin-surface-extrapolation-at-occlusion-boundaries,radar-camera-pixel-depth-association-for-depth-completion}, multi-view stereo~\cite{xu2021digging}, visual correspondence~\cite{truong2021learning}, face alignment~\cite{ luvli-face-alignment-estimating-landmarks-location-uncertainty-and-visibility-likelihood}, 3D reconstruction~\cite{2d-gans-meet-unsupervised-single-view-3d-reconstruction}, and face recognition~\cite{shi2019probabilistic,chang2020data,adaface-quality-adaptive-margin-for-face-recognition}.
In this work, we introduce an uncertainty solution in our dense correspondence model by representing $3$D points as distributions instead of deterministic points in our semantic part embedding space. Consequently, the learned variance of PEVs can be used as the measurement of the point-wise correspondence, which is suitable for generic objects with rich geometric and topological variations.  
%

%%%%%%%%%%%%%%%%%%%%%%%%%%%%%%%%%%%%%%%%%%%%%%
\subsection{Unsupervised Shape Co-Segmentation}
Co-segmentation is one of the fundamental tasks in geometry processing.
Prior works~\cite{yi2018deep,tulsiani2017learning,shu2016unsupervised} develop clustering strategies for meshes, given a handcrafted similarity metric induced by an embedding or graph~\cite{sidi2011unsupervised,xu2010style,huang2011joint}. 
The segmentation for each cluster is computed independently without accounting for statistics of shape variations, and the overall complexity of these methods is quadratic in the number of shapes in the collection.
Recently, BAE-NET~\cite{chen2019bae} presents an unsupervised branched autoencoder with $3$ fully-connected layers that discovers coarse segmentation of shapes by predicting implicit fields for each part. 
According to the evaluation in the BAE-NET~\cite{chen2019bae}, the $3$-layer network is the best choice for independent shape extraction, making it a suitable candidate for shape segmentation. 
However, the shallow network structure results in a limited shape representation power. 
In contrast, we extend the branched implicit function with a deep architecture, making it suitable for shape reconstruction as well.

\begin{figure}[t]
\centering    
\includegraphics[trim=0 0 0 0,clip, width=88mm]{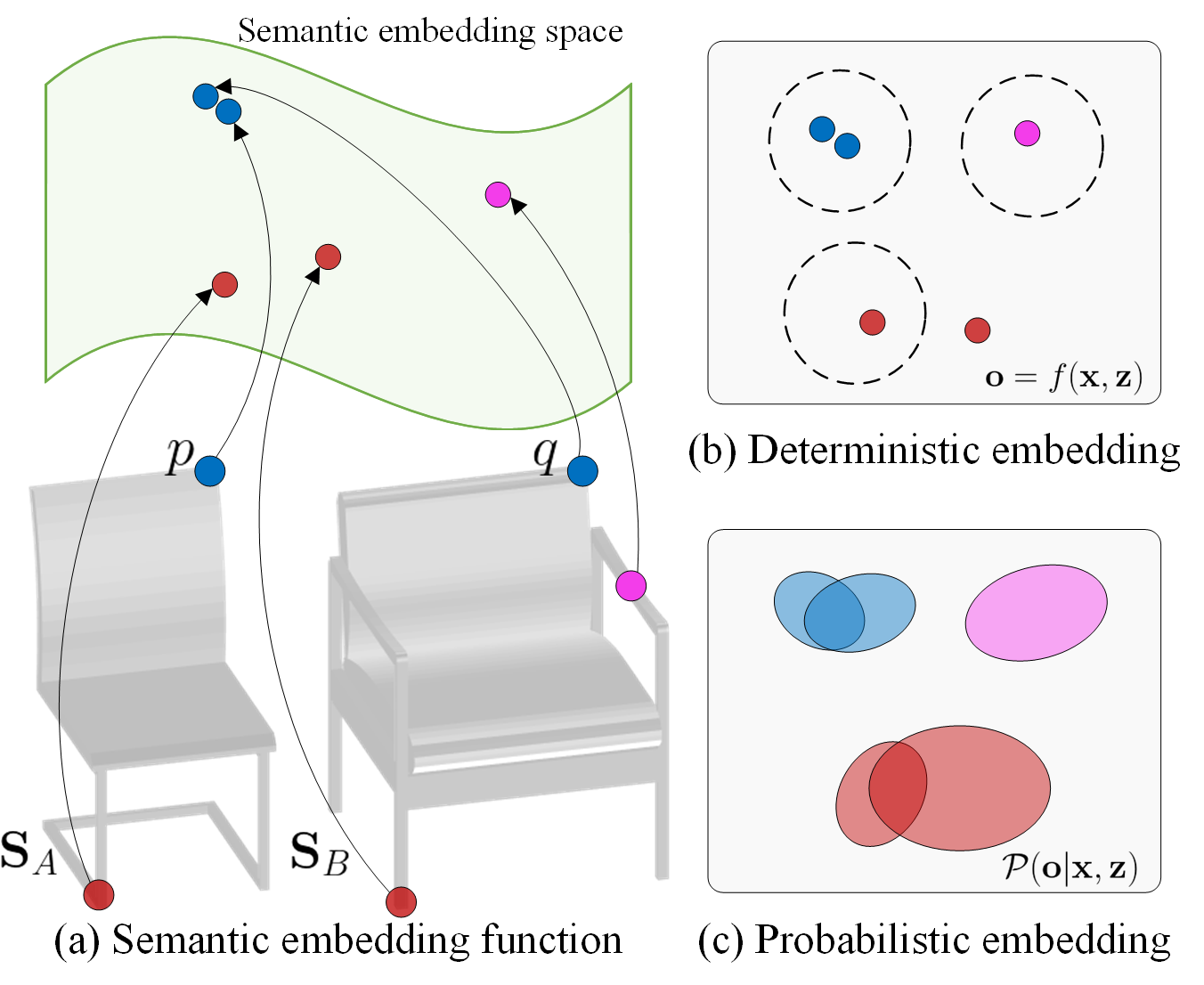} 
\caption{(a) We seek to learn a semantic embedding function, which maps a point from its 3D Euclidean space to the semantic embedding space. Consequently, when $p$ and $q$ locate in similar locations in the semantic embedding space, they have similar semantic meanings in their respective shapes. (b) In our preliminary work~\cite{liu2020correspondence}, the learned semantic embedding is a deterministic model, which represents each $3$D point as a \textbf{deterministic point} in the latent space without considering its feature ambiguity (\emph{i.e.}, the leg point of $\mathbf{S}_{A}$). (c) In this work, we propose to use probabilistic embeddings to give a \textbf{distributional} estimation of PEVs in the semantic space, which is able to capture a point-wise uncertainty in the dense correspondence model.}
\label{fig:uncertainty}
\end{figure}

%% file: sec_3_method.tex
%% -------------------
\begin{figure*}[t]
\centering    
\subfigure[]{\label{fig:flowchart_a}\includegraphics[height=47mm]{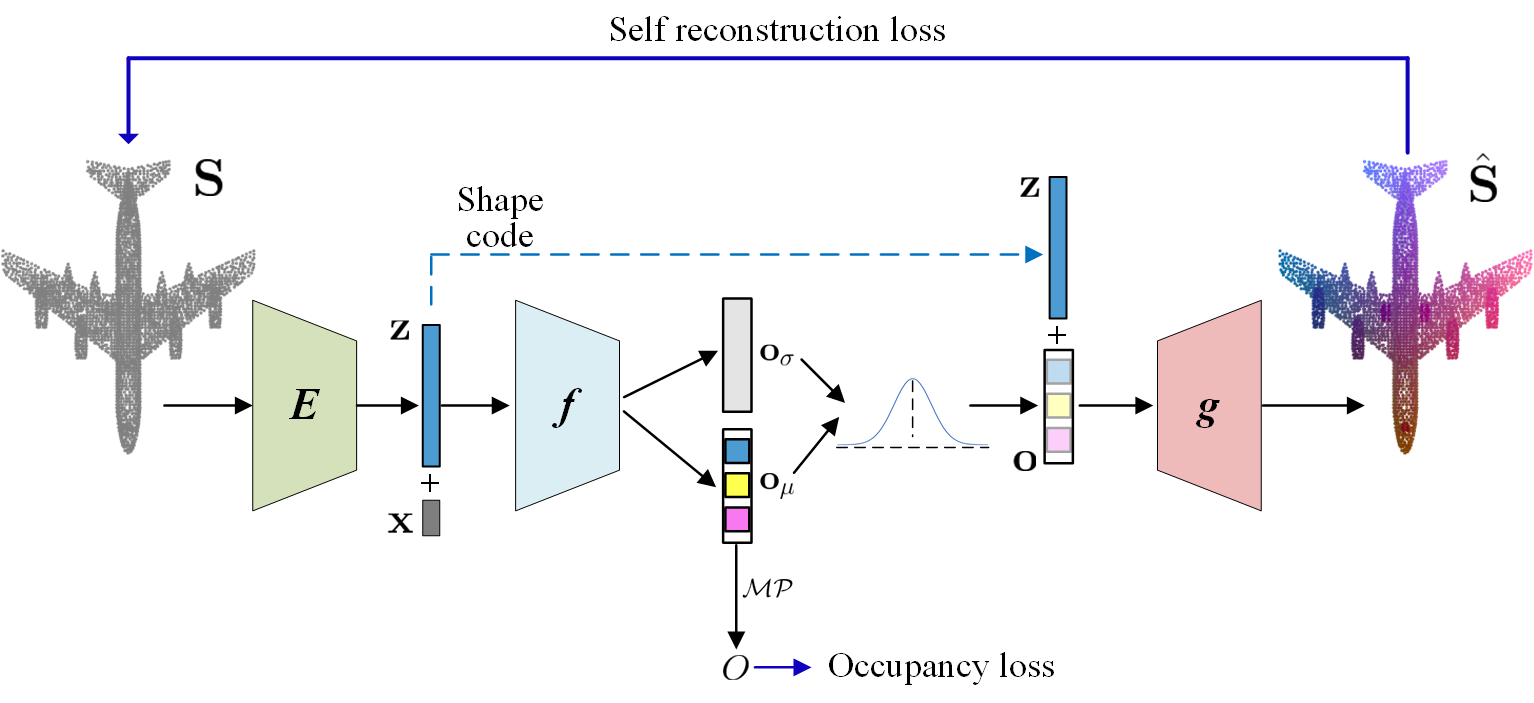}}
\hspace{2mm}
\subfigure[]{\label{fig:flowchart_b}\includegraphics[height=47mm]{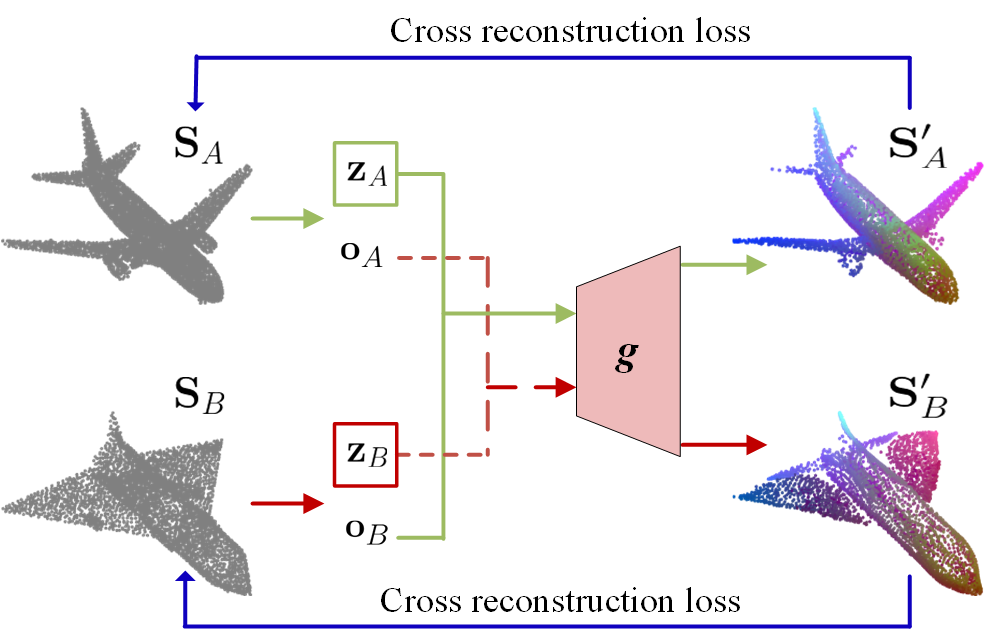}}
\caption{ \textbf{Model Overview.} (a) Given a shape $\mathbf{S}$, PointNet $E$ is used to extract the shape feature code $\mathbf{z}$. The parameters ($\mathbf{o}_{\mu}$, $\mathbf{o}_{\sigma}$) of the Gaussian distribution are predicted via a deep implicit function $f$. Then a stochastic part embedding vector $\mathbf{o}$ is sampled from $\mathcal{N}(\mathbf{o}; \mathbf{o}_{\mu},\mathbf{o}_{\sigma}^2\mathbf{I})$ in the semantic embedding space. We implement dense correspondence through an inverse  function mapping from $\mathbf{o}$ to recover the $3$D shape $\mathbf{\hat{S}}$. (b) To further make the learned part embedding consistent across all the shapes, we randomly select two shapes $\mathbf{S}_A$ and $\mathbf{S}_B$. By swapping the part embedding vectors, a cross-reconstruction loss is used to enforce the inverse function to recover to each other. $\mathcal{MP}$ denotes the max-pooling operator.}
\label{fig:flowchart}
\end{figure*}

\section{Proposed Method}\label{sec:method}
\subsection{Preliminaries}

%% -------------------
Let us first formulate the dense $3$D correspondence problem. 
Given a collection of $3$D shapes of the same object category, one may encode each shape $\mathbf{S}\in\mathbb{R}^{n\times 3}$ in a latent space $\mathbf{z}\in\mathbb{R}^{d} $.
As show in Fig.~\ref{fig:uncertainty}\textcolor{red}{(a)},
for any point $p\in \mathbf{S}_A$ in the source shape $\mathbf{S}_A$,  dense $3$D correspondence will find its semantic corresponding point $q\in \mathbf{S}_B$ in the target shape $\mathbf{S}_B$. if a semantic embedding function (SEF) $f: \mathbb{R}^3 \times \mathbb{R}^{d} \rightarrow \mathbb{R}^{k}$ is able to satisfy
\begin{equation}
\left(\min_{q\in \mathbf{S}_B}||f(p,\mathbf{z}_A)-f(q,\mathbf{z}_B)||_2\right)<\tau, \;\;\;\;\;     \forall p\in \mathbf{S}_A.
\label{eqn:obj}
\end{equation}
Here the SEF is responsible for mapping a point from its $3$D Euclidean space to the semantic embedding space.
When $p$ and $q$ have sufficiently similar locations in the semantic embedding space, they have similar semantic meaning, or functionality, in their respective shapes.
Hence $q$ is the corresponding point of $p$.
On the other hand, if their distance in the embedding space is too large ($\geq\tau$), there is no corresponding point in $ \mathbf{S}_B$ for $p$.
If SEF could be learned for a small $\tau$, the corresponded point $q$ of $p$ can be solved via $q=f^{-1}(f(p,\mathbf{z}_A), \mathbf{z}_B)$, where $f^{-1}(:,:)$ is the inverse function of $f$ that maps a point from the semantic embedding space back to the $3$D space.
Therefore, the dense $3$D correspondence problem amounts to learning the SEF and its inverse function. 

%% -------------------
\Paragraph{Probabilistic Semantic Embedding Learning}
Our preliminary work~\cite{liu2020correspondence} adopts a deterministic point representation for each $3$D point in the semantic embedding space. 
However, it is difficult to estimate an accurate point embedding for shape parts with semantic ambiguity, which usually has larger uncertainty in the embedding space (Fig.~\ref{fig:uncertainty}\textcolor{red}{(b)}). Also, these ambiguous features will negatively affect the mapping of the inverse function, leading a poor correspondence accuracy.  
To address this issue, we propose to utilize probabilistic embeddings to predict a distributional estimation $\mathcal{P}(\mathbf{o}|\mathbf{x},\mathbf{z})$ instead of a point estimation $f(\mathbf{x},\mathbf{z})$, for each $3$D point of the shapes in the semantic embedding space (Fig.~\ref{fig:uncertainty}\textcolor{red}{(c)}). Specifically, we define the PEV in the latent space as a Gaussian distribution:
\begin{equation}
\mathcal{P}(\mathbf{o}|\mathbf{x},\mathbf{z}) = \mathcal{N}(\mathbf{o}; \mathbf{o}_{\mu},\mathbf{o}_{\sigma}^2\mathbf{I}), 
\end{equation}
where the mean and variance of the Gaussian distribution are predicted by the function $f$:
$(\mathbf{o}_{\mu}, \mathbf{o}_{\sigma})=f(\mathbf{x},\mathbf{z})$.
Here,
The mean $\mathbf{o}_{\mu}$ can be regarded as the semantic feature of the point. The variance $\mathbf{o}_{\sigma}$ encodes the model's uncertainty along each feature dimension.

%% -------------------
\Paragraph{Implementation Solution}
As shown in Fig.~\ref{fig:flowchart}, we propose to leverage the topology-free implicit function, a conventional shape representation, to jointly serve as the SEF.
By assuming that corresponding points are similar in the embedding space, we explicitly implement an inverse function mapping from the embedding space to the $3$D space, so that the learning objectives can be more conveniently defined in the $3$D  space rather than the embedding space.
Both functions are jointly learned with an occupancy loss for accurate shape representation, and an uncertainty-aware self-reconstruction loss for the inverse function to recover itself. 
In addition, we propose an uncertainty-aware cross-reconstruction loss enforcing two objectives.
One is that the two functions can deform source shape points to be sufficiently close to the target shape.
The other is that the offset vectors between corresponding points, $\overrightarrow{pq}$, are {\it locally} smooth within the neighborhood of $p$.

%%%
\subsection{PointNet Encoder}
To perform dense correspondence for a $3$D shape, we need to first obtain a latent representation describing its overall shape. 
In this work, given a shape $\mathbf{S}\in\mathbb{R}^{n\times 3}$, we utilize a PointNet-based network to encode the shape into a latent code space. We adopt the original PointNet~\cite{qi2017pointnet} without the STN module to extract a global shape code $\mathbf{z}\in\mathbb{R}^{d}$:
\begin{equation}
E: \mathbb{R}^{n\times 3} \rightarrow \mathbb{R}^{d}.
\end{equation}

\subsection{Uncertainty-aware Implicit Function}~\label{sec:impl}
Based on the shape code $\mathbf{z}$ of an object, as in~\cite{chen2018learning,mescheder2018occupancy}, 
the $3$D shape of the object can be reconstructed by an implicit function. 
That is, given the $3$D coordinate of a query point $\mathbf{x}\in\mathbb{R}^{3}$, the implicit function 
assigns an occupancy probability $O$ between $0$ and $1$: $\mathbb{R}^3 \times \mathbb{R}^{d} \rightarrow [0,1]$, where $1$ indicates $\mathbf{x}$ is inside the shape, and $0$ outside.

This conventional function can not serve as our SEF, given its simple $1$D output.
Motivated by the unsupervised part segmentation~\cite{chen2019bae}, we adopt its branched layer as the  final layer of our implicit function, whose outputs are denoted by $\mathbf{o}_{\mu}\in\mathbb{R}^k$ and $\mathbf{o}_{\sigma}\in\mathbb{R}^k$: 
\begin{equation}
f: \mathbb{R}^3 \times \mathbb{R}^{d} \rightarrow (\mathbb{R}^{k}, \mathbb{R}^{k}).
\end{equation}
A max-pooling operator ($\mathcal{MP}$) leads to the final occupancy $O=\mathcal{MP}(\mathbf{o}_{\mu})$ by selecting one branch from $\mathbf{o}_{\mu}$, whose index indicates the unsupervisedly estimated part where $\mathbf{x}$ belongs to.
Conceptually, each element of $\mathbf{o}_{\mu}$ shall indicate the occupancy value of $\mathbf{x}$ w.r.t.~the respective part.
Since $\mathbf{o}$ appears to represent the occupancy of $\mathbf{x}$ w.r.t.~all semantic parts of the object, the latent space of $(\mathbf{o}_{\mu}, \mathbf{o}_{\sigma})$ can be the desirable probabilistic semantic embedding.
In our implementation, $f$ is a $3$-layer multilayer perceptron (MLP).  
The final layer consists of two separate fully connected (FC) layers designed to produce the mean $\mathbf{o}_{\mu}$ and variance $\mathbf{o}_{\sigma}$ of the Gaussian distribution.

\begin{figure}[t]
\centering    
\includegraphics[trim=6mm 0 6mm 0,clip, width=88mm]{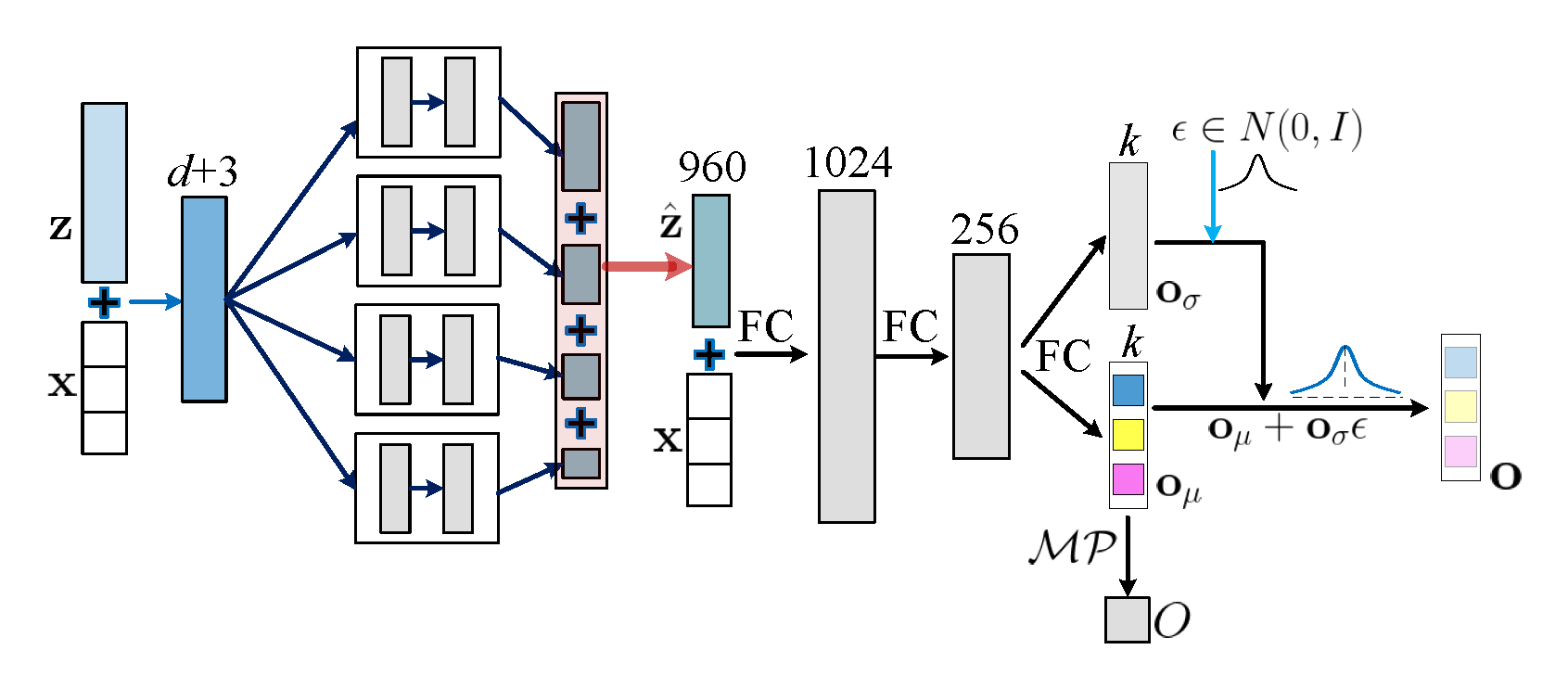} 
\caption{Uncertainty-aware deep branched implicit function network. Four branched MLPs are first utilized to produce point features in different scales. Then the features are aggregated into a single point-wise latent vector $\hat{\mathbf{z}}$. The final $4$ fully connected layers predict the mean $\mathbf{o}_{\mu}{\in}\mathbb{R}^{k}$ and variance $\mathbf{o}_{\sigma}{\in}\mathbb{R}^{k}$ of the Gaussian distribution. The part embedding vector $\mathbf{o}$ of each point $\mathbf{x}$ is not a deterministic point embedding any point, but a stochastic embedding sampled from $\mathcal{N}(\mathbf{o}; \mathbf{o}_{\mu},\mathbf{o}_{\sigma}^2\mathbf{I})$. A max-pooling operator leads to the final occupancy $O{=}\mathcal{MP}(\mathbf{o}_{\mu})$.}
\label{fig:implicit_hier}
\end{figure}

\Paragraph{Deep Branched Implicit Function}
As detailed in the work~\cite{chen2019bae}, the network structure of $3$-layer implicit function can be sensitive to the initial parameters and it cannot infer semantic information when the number of layers is greater than $3$. 
As a result, the limited depth of the network makes it difficult to represent shape details, as well as obtain fine-grained semantic correspondence. %consistency and shape representation. 
To address these issues, we introduce a \emph{\textbf{deep}} branched implicit function network. Specifically, as shown in Fig.~\ref{fig:implicit_hier}, we first generate deep point-specific features via four parallel MLPs. 
We then concatenate those features to produce a point-specific latent code $\hat{\mathbf{z}}$. The final $3$ FC layers produce the occupancy value of the query point. 
Querying deep features extracted at continuous $3$D locations used in implicit function learning allows us to reconstruct the local geometric structure of generic objects, enhancing the point-wise semantic representation in the semantic embedding.

%% ----------------
%% ----------------

%%
\subsection{Inverse Implicit Function}
Given the objective function in Eqn.~\ref{eqn:obj}, one may consider that learning SEF, $f$, would be sufficient for dense correspondence.
However, there are two problems with this. 
First of all, to find correspondence of $p$, we need to compute $f^{-1}(f(p,\mathbf{z}_A), \mathbf{z}_B)$, \emph{i.e.}, assuming the output of  $f(q,\mathbf{z}_B)$ equals $f(p,\mathbf{z}_A)$ and solve for $q$ via iterative back-propagation. 
This could be computationally inefficient during inference.
Secondly, it is easier to define shape-related constraints or loss functions between $f^{-1}(f(p,\mathbf{z}_A), \mathbf{z}_B)$ and $q$ in the $3$D space,  rather than those between $f(q,\mathbf{z}_B)$ and $f(p,\mathbf{z}_A)$ in the embedding space. 

To this end, we define the inverse implicit function to take the probabilistic part embedding vector (PEV) $\mathbf{o}$ and the shape code $\mathbf{z}$ as inputs, and recover the corresponding $3$D location:
\begin{equation}
g: \mathbb{R}^k \times \mathbb{R}^{d} \rightarrow \mathbb{R}^3.
\end{equation}

The probabilistic PEV can be reformulated as $\mathbf{o}=\mathbf{o}_{\mu}+\epsilon\mathbf{o}_{\sigma}$, where $\epsilon\sim\mathcal{N}(0,\mathbf{I})$. 
We also use an MLP network to implement $g$. 
With $g$, we can efficiently compute $g(f(p,\mathbf{z}_A), \mathbf{z}_B)$ via forward passing, without iterative back-propagation.

%%%
\subsection{Training Loss Functions}
We jointly train our implicit function and inverse function by minimizing three losses: the occupancy loss $\mathcal{L}^{occ}$, the uncertainty-aware self-reconstruction loss $\mathcal{L}^{SR}$, and the uncertainty-aware cross-reconstruction loss $\mathcal{L}^{CR}$, \emph{i.e.},
\begin{equation}
\mathcal{L}^{all} = \mathcal{L}^{occ} + \mathcal{L}^{SR} + \mathcal{L}^{CR},
\label{eqn:loss}
\end{equation}
where $\mathcal{L}^{occ}$ measures how accurately $f$ predicts the occupancy of the shapes,  $\mathcal{L}^{SR}$ enforces that $g$ is an inverse function of $f$, 
and $\mathcal{L}^{CR}$ strives for part embedding consistency across all shapes in the collection. 
We first explain how we prepare the training data, and then provide the details of the loss functions.

\subsubsection{Training Samples}
Given a collection of $N$ raw $3$D surfaces $\{\mathbf{S}^{raw}_{i}\}_{i=1}^{N}$ with a consistent upright orientation, we first normalize the raw surfaces by uniformly scaling the object such that the diagonal of its tight bounding box has a constant length and make the surfaces watertight by converting them to voxels. 
In order to train the implicit function model, following the sample scheme of~\cite{chen2018learning}, we randomly sample and obtain $K$ spatial points $\{\mathbf{x}_{j}\}_{j=1}^{K}$ and their occupancy labels $\{\tilde{O}_{j}\}_{j=1}^{K}\in\{0,1\}$ near the surface, which are $1$ for the inside points and $0$ otherwise. In addition, to learn discriminative shape codes, we further uniformly sample $n$ {\it surface points} to represent $3$D shapes, resulting in $\{\mathbf{S}_{i}\}_{i=1}^{n}$. 

\subsubsection{Occupancy Loss} This is a $L_2$ error between the label and estimated occupancy of all shapes:
\begin{equation}
\mathcal{L}^{occ} = \sum_{i=1}^{N}  \sum_{j=1}^{K} \| \mathcal{MP}(f_{\mathbf{o}_{\mu}}(\mathbf{x}_{j}, \mathbf{z}_{i})) - \tilde{O}_{j}\|^{2}_{2}.
\label{eqn:loss_point}
\end{equation}

\subsubsection{Uncertainty-aware Self-Reconstruction Loss}
 the inverse function aims to map from the embedding space to the $3$D space. The variance could actually be regarded as the uncertainty measuring the confidence of the inverse mapping. Following the minimisation objective suggested by~\cite{kendall2017uncertainties}, we supervise the inverse function by recovering input surface $\mathbf{S}_{i}$:
\begin{equation}
\begin{split}
\mathcal{L}^{SR} = & \sum_{i=1}^{N}  \sum_{j=1}^{n} \frac{1}{2}(\mathbf{o}_{\sigma}^{(j)})^{-2}\| g(f(\mathbf{S}_i^{(j)}, \mathbf{z}_{i} ),\mathbf{z}_{i} ) - \mathbf{S}_i^{(j)}\|^{2}_{2} \\
 & + \frac{1}{2}\log(\mathbf{o}_{\sigma}^{(j)})^{2},
\label{eqn:loss_self}
\end{split}
\end{equation}
where $\mathbf{S}_i^{(j)}$ is the $j$-th point of shape $\mathbf{S}_i$ and $\mathbf{o}_{\sigma}^{(j)}$ denotes the mean value across all dimensions of its variance.
The first term $\frac{1}{2}(\mathbf{o}_{\sigma}^{(j)})^{-2}$ serves as a weighted distance which assigns larger weights to less uncertainty vectors. The second term $\frac{1}{2}\log(\mathbf{o}_{\sigma}^{(j)})^{2}$ penalizes points with high uncertainties. In practice, we train the function $f$ to predict the log variance $\log\mathbf{o}_{\sigma}^2$ for stable optimization.

\subsubsection{Uncertainty-aware Cross-Reconstruction Loss} The cross-reconstruction loss is designed to encourage the resultant PEVs to be similar for densely corresponded points from any two shapes.
As shown in Fig.~\ref{fig:flowchart}, from a shape collection we first randomly select two shapes $\mathbf{S}_{A}$ and $\mathbf{S}_{B}$. 
The implicit function $f$ generates PEV sets $\{\mathbf{o}_{A}\}$ ($\{\mathbf{o}_{B}\}$), given $\mathbf{S}_{A}$ ($\mathbf{S}_{B}$) and their respective shape codes $\mathbf{z}_{A}$ ($\mathbf{z}_{B}$) as inputs. 
Then we swap their PEVs and feed the concatenated vectors to the inverse function $g$: $\mathbf{S}_{A}^{'(j)} = g(\mathbf{o}_{B}^{(j)}$, $\mathbf{z}_{A}),  \mathbf{S}_{B}^{'(j)} = g(\mathbf{o}_{A}^{(j)}, \mathbf{z}_{B})$.   
If the part embedding is point-to-point consistent across all shapes, the inverse function should recover each other, \emph{i.e.}, $\mathbf{S}'_{A}\approx \mathbf{S}_{A}$, $\mathbf{S}'_{B}\approx \mathbf{S}_{B}$. 
Towards this goal, we exploit several loss functions to minimize the pairwise difference for each of these two shape pairs:
\begin{equation}
\mathcal{L}^{CR} = \lambda_{1}\mathcal{L}^{CD} + \lambda_{2}\mathcal{L}^{EMD} + \lambda_{3}\mathcal{L}^{nor} + \lambda_{4}\mathcal{L}^{smo}, 
\label{eqn:const}
\end{equation}
where $\mathcal{L}^{CD}$ is the Chamfer Distance (CD) loss, $\mathcal{L}^{EMD}$ the Earth Mover distance (EMD) loss, $\mathcal{L}^{nor}$ the surface normal loss, $\mathcal{L}^{smo}$  the smooth correspondence loss, and $\lambda_{i}$ are the weights.  
The first three terms focus on shape similarity, while the last one encourages the correspondence offsets to be locally smooth. We empirically apply uncertainty learning in  $\mathcal{L}^{CD}$ only since $\mathcal{L}^{CD}$ essentially reflects the main results of the cross-reconstruction process.

\Paragraph{Uncertainty-aware Chamfer Distance Loss} is defined as:
\begin{equation}
\mathcal{L}^{CD} = d_{CD}(\mathbf{S}_{A}, \mathbf{S}'_{A})+d_{CD}(\mathbf{S}_{B}, \mathbf{S}'_{B}),
\label{eqn:cd_loss}
\end{equation}
where CD is calculated as~\cite{qi2017pointnet}:
\begin{equation}
\begin{split}
d_{CD}(\mathbf{S}, \mathbf{S}') = & \frac{1}{2}(\mathbf{o}_{\sigma}^{(p)})^{-2} \sum_{p\in\mathbf{S}}\min_{q\in\mathbf{S}'}\| p-q\|^{2}_{2}+ \frac{1}{2}log(\mathbf{o}_{\sigma}^{(p)})^{2} \\
& +\frac{1}{2}(\mathbf{o}_{\sigma}^{(p)})^{-2}\sum_{q\in\mathbf{S}'}\min_{p\in\mathbf{S}}\| p-q\|^{2}_{2}+ \frac{1}{2}log(\mathbf{o}_{\sigma}^{(p)})^{2},
\label{eqn:cd}
\end{split}
\end{equation}
where $\mathbf{o}_{\sigma}^{(p)}$ is the mean of all values in variance of point $p$. Similarly, the variance in the semantic embedding learns the correspondence uncertainty through such cross-reconstruction.

%--------------------------
 \begin{algorithm}[t]
 \setstretch{1.15}
 \caption{Dense correspondence inference.}
 \begin{algorithmic}[1]
 \renewcommand{\algorithmicrequire}{\textbf{Input:}} 
 \renewcommand{\algorithmicensure}{\textbf{Output:}}
 \REQUIRE Two surface point sets: $\mathbf{S}_A$ (Source) and $\mathbf{S}_B$ (Target). 
 \ENSURE The corresponding point sets and their confidence scores $\mathbf{S}_{A\to B}=\{q,C\}$ on $\mathbf{S}_B$.
 \\ \textit{Initialisation} :
  \STATE $\mathbf{z}_{A} \leftarrow E(\mathbf{S}_A)$, $\mathbf{z}_{B} \leftarrow E(\mathbf{S}_B)$;
  \STATE $\{\mathbf{o}_{\mu B}\} \leftarrow,  f(\mathbf{z}_{B},\mathbf{S}_B)$, $\{\mathbf{o}_{\mu A}\} \leftarrow f(\mathbf{z}_{A},\mathbf{S}_A)$;
   \STATE $\mathbf{S}'_A \leftarrow g(\mathbf{z}_{A},\{\mathbf{o}_{\mu B}\})$;
   \STATE $\mathbf{S}_{A\to B} \gets \emptyset$
 \\ \textit{LOOP Search Function:}
  \FOR {each point $p$ in $\mathbf{S}_A$}
  \STATE Find a preliminary correspondence $q'$ in $\mathbf{S}'_A$ via $q'=\arg \min_{q'\in\mathbf{S}'_A}\| p - q'\|_{2}$;
  \STATE Knowing the index of $q'$ in $\mathbf{S}'_A$, the same index in  $\mathbf{S}_{B}$ refers to the final correspondence $q\in\mathbf{S}_{B}$;
  \STATE Compute the confidence score via Eqn.~\ref{eqn:confidence_score}.
  
  \IF {$\mathcal{C} > \tau'$}
  \STATE $\mathbf{S}_{A\to B} \gets \mathbf{S}_{A\to B}\, ||\,  (q, C)$; 
  \ELSE
  \STATE $\mathbf{S}_{A\to B} \gets \mathbf{S}_{A\to B}\, ||\,  (\emptyset, C)$; 
  \ENDIF
  \ENDFOR
  \RETURN $\mathbf{S}_{A\to B}$
 \end{algorithmic} 
  \label{alg:inference}
 \end{algorithm}
%--------------------------

%%
\Paragraph{Earth Mover Distance Loss} is defined as:
\begin{equation}
\mathcal{L}^{EMD} = d_{EMD}(\mathbf{S}_{A}, \mathbf{S}'_{A})+d_{EMD}(\mathbf{S}_{B}, \mathbf{S}'_{B}),
\label{eqn:emd_loss}
\end{equation}
where EMD is the minimum of the sum of distances between a point in one set and a point in another set over all possible permutations of correspondences~\cite{qi2017pointnet}:
\begin{equation}
d_{EMD}(\mathbf{S}, \mathbf{S'}) =  \min_{\Phi:\mathbf{S}\rightarrow \mathbf{S}'}\sum_{p\in\mathbf{S}}\| p-\Phi(p) \|_{2},
\label{eqn:emd}
\end{equation}
where $\Phi$ is a bijective mapping. 

\Paragraph{Surface Normal Loss} An appealing property of implicit representation is that the surface normal can be analytically computed using the spatial derivative $\frac{ \partial \mathcal{MP}(f(\mathbf{x}, \mathbf{z}))}{\partial\mathbf{x}}$ via back-propagation through the network. Hence, we are able to define the surface normal distance on the point sets. 
\begin{equation}
\mathcal{L}^{nor}= d_{nor}(\mathbf{n}_{A}, \mathbf{n}'_{A})+
d_{nor}(\mathbf{n}_{B}, \mathbf{n}'_{B}),
\label{eqn:normal_loss}
\end{equation}
where $\mathbf{n}_{*}$ is the surface normal of $\mathbf{S}_{*}$. We measure $d_{nor}$ by the Cosine similarity distance:
\begin{equation}
d_{nor}(\mathbf{n},\mathbf{n}')=\frac{1}{n}\sum_{i}(1-\mathbf{n}_i\cdot\mathbf{n}'_{i}),
\end{equation}
where $\cdot$ denotes the dot-product.

\Paragraph{Smooth Correspondence Loss} encourages that the correspondence offset vectors $\Delta \mathbf{S}_{AB}= \mathbf{S}'_{B}-\mathbf{S}_{A}$, $\Delta \mathbf{S}_{BA}= \mathbf{S}'_{A}-\mathbf{S}_{B}$ of neighboring points are as similar as possible to ensure a smooth deformation:
\begin{equation}
\begin{split}
\mathcal{L}^{smo}= \sum_{a\in\mathbf{S}_{A},a'\in\mathbb{N}(a)}\| \Delta \mathbf{S}_{AB}^{(a)}-\Delta \mathbf{S}_{AB}^{(a')}\|_{2} + \\ \sum_{b\in\mathbf{S}_{B},b'\in\mathbb{N}(b)}\| \Delta \mathbf{S}_{BA}^{(b)}-\Delta \mathbf{S}_{BA}^{(b')}\|_{2},
\label{eqn:smooth}
\end{split}
\end{equation}
where  $\mathbb{N}(a)$ and $\mathbb{N}(b)$ are neighborhoods for $a$ and $b$ respectively. Here, for the local neighborhood selection, we utilize the radius-based ball query strategy~\cite{qi2017pointnet++} with the radius being $0.1$.

%%%
\subsection{Inference}
During inference, our method can offer both shape segmentation and dense correspondence for $3$D shapes. 
As each element of PEV learns a compact representation for one common part of the shape collection, the shape segmentation of $p$ is the index of the element being max-pooled from its PEV.
As both the implicit function $f$ and its inverse $g$ are point-based, the number of input points to $f$ can be arbitrary during inference.
As depicted in Algorithm~\ref{alg:inference}, given two point sets $\mathbf{S}_A$, $\mathbf{S}_B$ with shape codes $\mathbf{z}_A$ and $\mathbf{z}_{B}$, $f$ generates the mean of PEVs $\mathbf{o}_{\mu A}$ and $\mathbf{o}_{\mu B}$, and $g$ outputs cross-reconstructed shape $\mathbf{S}'_A$. 
For any query point $p\in \mathbf{S}_A$, a preliminary correspondence may be found by the nearest neighbor search in $\mathbf{S}'_A$: $q'=\arg \min_{q'\in\mathbf{S}'_A}\| p - q'\|_{2}$. 
Knowing the index of $q'$ in $\mathbf{S}'_A$, the same index in $\mathbf{S}_{B}$ refers to the final correspondence $q\in\mathbf{S}_{B}$.

Finally, given the probabilistic semantic embedding $(\mathbf{o}_{A}^{i_p},\mathbf{o}_{ B}^{i_q})$ of the corresponding points ($p$,$q$), the correspondence confidence can be computed by measuring the ``likelihood'' of them:$\mathcal{P}(\mathbf{o}_{ A}^{i_p}=\mathbf{o}_{ B}^{i_q})$, where $\mathbf{o}_{ A}^{i_p} \sim \mathcal{P}(\mathbf{o}|p,\mathbf{z}_A)$ and $\mathbf{o}_{ B}^{i_q} \sim \mathcal{P}(\mathbf{o}|q,\mathbf{z}_B)$.
In practice, we adopt the mutual likelihood score as the confidence score~\cite{shi2019probabilistic}:
\begin{equation}
\footnotesize
    \mathcal{C}= - \sum_{l}^{k} \left ( \frac{({\mathbf{o}_{\mu A}^{i_p}}^{(l)}-{\mathbf{o}_{\mu B}^{i_q}}^{(l)})}{(\mathbf{o}_{\sigma A}^{i_p})^{2(l)}+(\mathbf{o}_{\sigma B}^{i_q})^{2(l)}} + \log\left ((\mathbf{o}_{\sigma A}^{i_p})^{2(l)}+(\mathbf{o}_{\sigma B}^{i_q})^{2(l)}\right)\right )
    \label{eqn:confidence_score}
\end{equation}
where $i_p$ is the index of $p$ in $\mathbf{S}_{A}$. ${\mathbf{o}_{\mu}}^{(l)}$ refers to the $l^{th}$ dimension of $\mathbf{o}_{\mu}$ and similarly for ${\mathbf{o}_{\sigma}}^{(l)}$.
Here, $\mathcal{C}$ is normalized to the range of $[0,1]$ with min-max normalization for all the testing samples.
Since the learned part embedding is discriminative among different parts of a shape, the distance of PEVs is suitable to define confidence.
When $\mathcal{C}$ is larger than a pre-defined threshold $\tau'$, this $p\rightarrow q$ correspondence is valid; otherwise $p$ has no correspondence in $\mathbf{S}_{B}$.

% -------------Figures of experiment-----------------  
\begin{figure*}[t]
    \centering
    \resizebox{1\linewidth}{!}{
    \includegraphics[trim={0.5cm 0 0.5cm 0},clip, width=\linewidth]{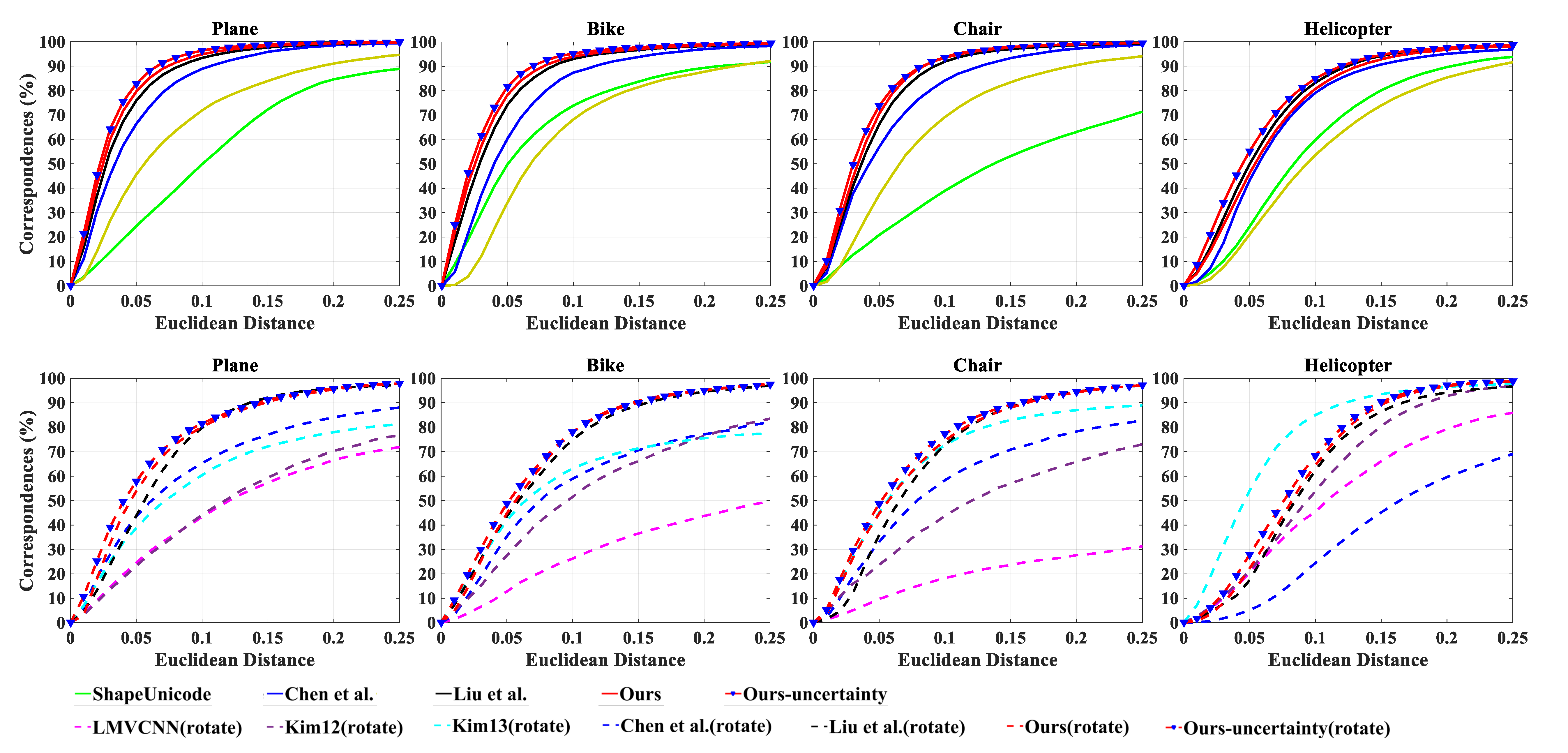}
    }
    \vspace{-2mm}
    \caption{ Correspondence accuracy for $4$ categories in the BHCP benchmark. The solid and dashed lines are for the aligned (top) and unaligned (bottom) setting respectively. All baseline results are quoted from~\cite{chen2020unsupervised,kim2013learning}.}
    \label{fig:semantic}
\end{figure*}

\begin{figure*}[t]
\centering    
\subfigure[]{\label{fig:corr_example_a}\includegraphics[trim=0 2 0 2,clip, height=70mm]{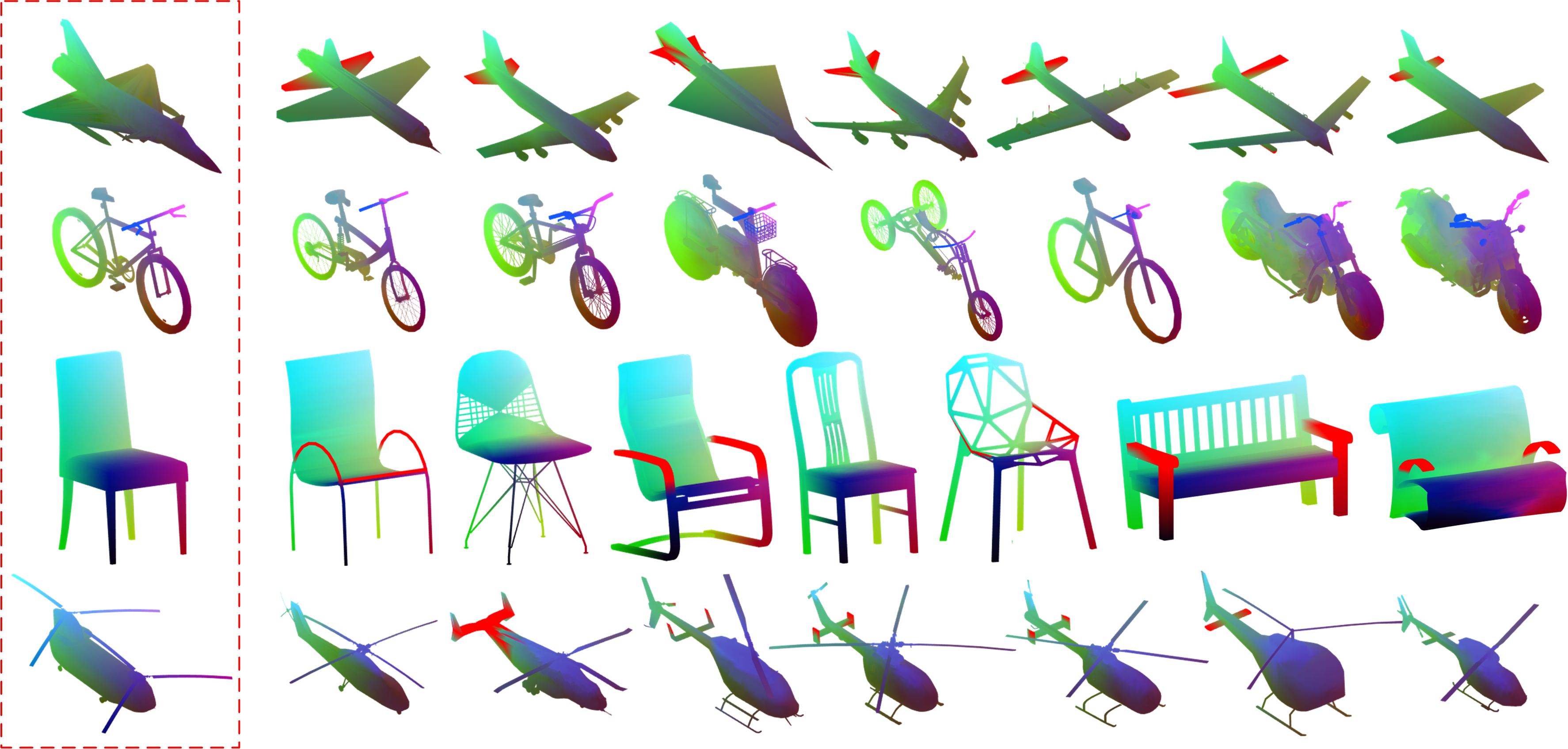}} 
\hspace{1mm}
\subfigure[]{\label{fig:corr_example_b}\includegraphics[trim=0 2 0 2,clip,height=70mm]{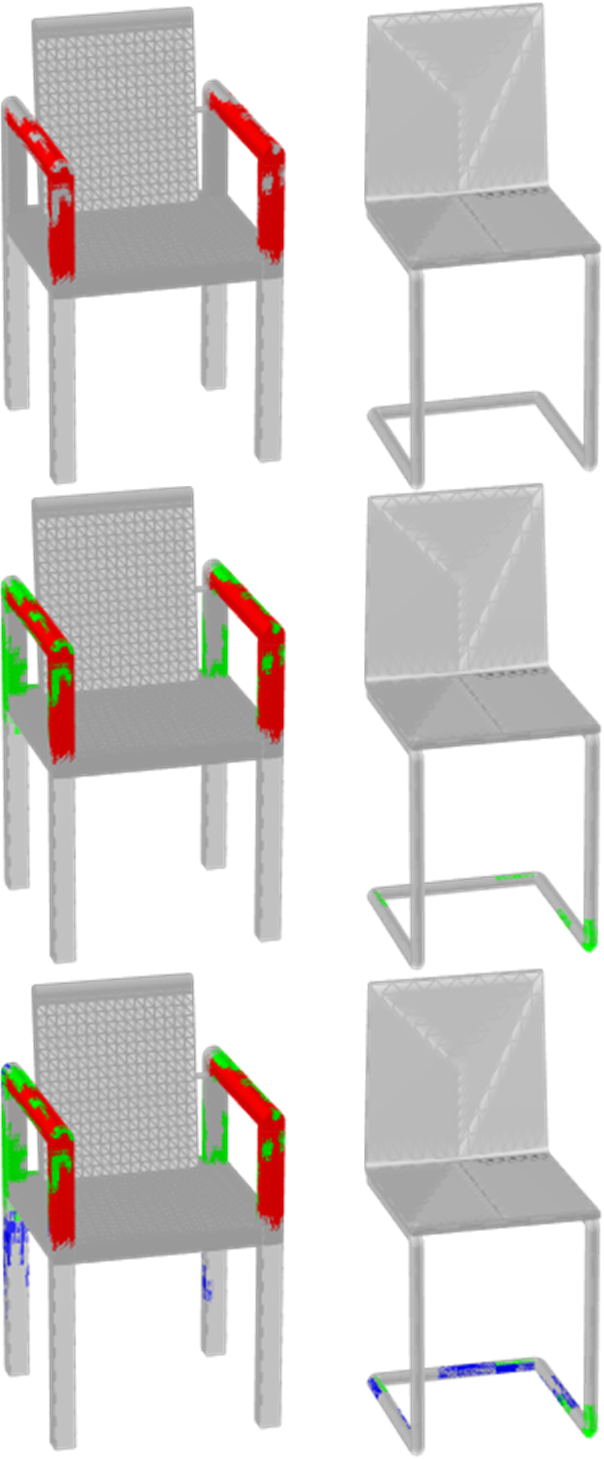}}
\vspace{-2mm}
\caption{(a) Dense correspondences in $4$ categories. Each row shows one target shape $\mathbf{S}_{B}$ (red box) and its pair-wise corresponded $6$ source shapes $\mathbf{S}_{A}$. 
Given a spatially colored $\mathbf{S}_{B}$, the $p\rightarrow q$ correspondence enables to assign  $p\in\mathbf{S}_{A}$ with the color of $q\in\mathbf{S}_{B}$, or with red if $q$ is non-existing.
(b) For one pair of shapes, the non-existence correspondences are impacted by the confidence threshold $\tau'$. The colored regions progressively show the non-existence correspondences between the two shapes where the confidence score $\mathcal{C}$ is in the range of $[0, 0.3]$ (red), $(0.3, 0.5]$ (green), and $(0.5,0.7]$ (blue).
}
\label{fig:corr_example}
\end{figure*}

% -------------Figures of experiment----------------- 

\subsection{Implementation Detail}
\subsubsection{Sampling Point-Value Pairs}
The training of implicit function network needs point-value pairs.
Following the sampling strategy of~\cite{chen2018learning}, we obtain the paired data $\{\mathbf{x}_{j}, \tilde{O}_{j}\}_{j=1}^{K}$ offline. $\mathbf{x}_{j},\tilde{O}_{j}$ are the spatial point and its corresponding occupancy label. 
For each $3$D shape, we utilize the technique of Hierarchical Surface Prediction (HSP)~\cite{hane2017hierarchical} to generate the voxel models at different resolutions ($16^3$, $32^3$, $64^3$). We then respectively sample points ($K=4,096$, $K=8,192$, $K=32,768$) on three resolutions in order to train the implicit function {\it progressively}.

\begin{table}[t]
\centering 
\renewcommand\arraystretch{1.1}
\caption{Three stages of the training process.}  
\begin{tabular}{l| c| c}
\toprule
 Stage & Updated networks   & Loss functions\\ 
\midrule 
\midrule
 $1$ & $E$, $f$       &    $\mathcal{L}^{occ}$\\ \hline
 $2$ & $E$, $f$, $g$  &    $\mathcal{L}^{occ}$ and $\mathcal{L}^{SR}$\\ \hline
 $3$ & $E$, $f$, $g$  &    $\mathcal{L}^{all}$\\
\bottomrule
\end{tabular}
\label{tab:train_process}
\end{table}

\subsubsection{Training Process} 
We summarize the training process in Tab.~\ref{tab:train_process}. 
In order to speed up the training process, our method is trained in three stages:
$1$) To encode the shape codes of the input shapes, PointNet $E$ and implicit function $f$ are first trained on sampled point-value pairs via Eqn.~\ref{eqn:loss_point}; 
$2$) To enforce inverse implicit function $g$ to recover $3$D points from PEVs,
$E$, $f$, and inverse function $g$ are jointly trained via  Eqn.~\ref{eqn:loss_point} and~\ref{eqn:loss_self}; 
$3$) To further facilitate the learned PEVs to be consistent for densely corresponded points from any two shapes, we jointly train $E$, $f$, and $g$ with $\mathcal{L}^{all}$. 
In Stage $1$, we adopt a progressive training technique~\cite{chen2018learning} to train our implicit function on data with gradually increasing resolutions ($16^3{\rightarrow}32^3{\rightarrow}64^3$), which stabilizes and significantly speeds up the training process. 

In experiments, we set $n=8{,}192$, $d=256$, $k=12$, $\tau'=0.2$, $\lambda_1=10$, $\lambda_2=1$, $\lambda_3=0.01$, $\lambda_4=0.1$. 
We implement our model in Pytorch and use
Adam optimizer at a learning rate of $0.0001$ in all three stages.

%% file: sec_4_exp.tex
\section{Experiments}\label{sec:exp}

%%%
\subsection{3D Semantic Correspondence}\label{sec:exp_correspondence}
\Paragraph{Data} 
We evaluate our proposed algorithm on the task of $3$D semantic point correspondence, a special case of dense correspondence, with two motivations: 1) no database of man-made objects has ground-truth dense correspondence; and 2) there is far less prior work in dense correspondence for man-made objects than the semantic correspondence task, which has strong baselines for comparison.
Thus, to evaluate semantic correspondence, we train on ShapeNet~\cite{chang2015shapenet} and test on BHCP~\cite{kim2013learning} following the experimental protocol of~\cite{huang2017learning,chen2020unsupervised}. 
For training, we use a subset of ShapeNet including plane ($500$), bike ($202$), and chair ($500$) categories to train $3$ individual models. 
For testing, BHCP provides ground-truth semantic points ($7$-$13$ per shape) of $404$ shapes including plane ($104$), bike ($100$), chair ($100$), and helicopter ($100$). 
We generate all pairs of shapes for testing, \emph{i.e.}, $9,900$ pairs for bikes.
The helicopter category is tested with the plane model as~\cite{huang2017learning,chen2020unsupervised} did. 
As BHCP shapes are with rotations, prior works test on either one or both settings: aligned and unaligned, {\it i.e.}, $0^\circ$ vs.~arbitrary relative pose of two shapes. 
We evaluate both settings.

%% -------------------------------------------------------------

%-----
\Paragraph{Baseline} We compare our work with multiple state-of-the-art (SoTA) baselines. 
Kim12~\cite{kim2012exploring} and Kim13~\cite{kim2013learning} are traditional optimization methods that require part labels for templates and employ collection-wise co-analysis. 
LMVCNN~\cite{huang2017learning}, ShapeUnicode~\cite{muralikrishnan2019shape}, AtlasNet2~\cite{deprelle2019learning}, Chen \emph{et al.}~\cite{chen2020unsupervised} and Liu~\emph{et al.}~\cite{liu2020correspondence} are all learning based, where~\cite{huang2017learning} require ground-truth correspondence labels for training. Despite \cite{chen2020unsupervised} only estimates a fixed number of sparse points, ~\cite{chen2020unsupervised} and ours are trained \textbf{without} labels. 
As optimization-based methods and~\cite{huang2017learning} are designed for the unaligned setting, we also train a rotation-invariant version of ours by supervising $E$ to predict additional rotation parameters, which is applied to rotate the input query point before feeding the point to $f$. In addition, we report results from two models: without uncertainty learning (\textbf{Ours}) and with uncertainty learning (\textbf{Ours-uncertainty}).

%-----
\Paragraph{Results} The correspondence accuracy is measured by the fraction of correspondences whose error is below a given threshold of Euclidean distances. 
As in Fig.~\ref{fig:semantic}, the solid lines show the results on the aligned data and dotted lines on the unaligned data. We can clearly observe that our method outperforms baselines in the plane, bike, and chair categories on aligned data. Note that Kim13~\cite{kim2013learning} has slightly higher accuracy than ours on the helicopter category, likely due to the fact that ~\cite{kim2013learning} tests with the \emph{helicopter-specific model}, while we test on the {\it unseen} helicopter category with a plane-specific model.
At the distance threshold of $0.05$, compared to our preliminary work~\cite{liu2020correspondence}, the \textbf{Ours} setting improves on average $5.3\%$ accuracy relatively in $3$ (Plane, Bike, and Chair) categories. While the \textbf{Ours-uncertainty} setting improves on average $9.2\%$ accuracy, and achieves $9.6\%$ improvement in the unseen Helicopter category. 
Moreover, compared to the best-performing SoTA baseline~\cite{chen2020unsupervised}, our average relative improvement is $29.3\%$ in $4$ categories.
We can clearly observe that both proposed deep branched implicit function and uncertainty learning can significantly improve the performance of semantic correspondence. \emph{In the rest of experiments, we will use the model trained with \textbf{Ours-uncertainty} setting (unless specified otherwise)}.

For unaligned data, both two settings achieve competitive performance as baselines.
While it has the best AUC overall, it is worse at the threshold between $[0,0.05]$.
The main reason is the implicit network itself is sensitive to rotation.
Note that this comparison shall be viewed in the context that most baselines use extra cues during training or inference, as well as the high inference speed of our learning-based approach.
For example, Kim$13$ requires a part-based template during inference.

Some visual results of dense correspondences are shown in Fig.~\ref{fig:corr_example_a}.
Note the amount of non-existent correspondence is impacted by the threshold $\tau'$ as in Fig.~\ref{fig:corr_example_b}. 
A larger $\tau'$ discovers more subtle non-existence correspondences.
This is expected as the division of semantically corresponded or not can be blurred for some shape parts.

\begin{figure}[t]
\centering     
\subfigure[]{\label{fig:ablation_study1_noise}\includegraphics[trim=2 4 4 2,clip, height=34mm]{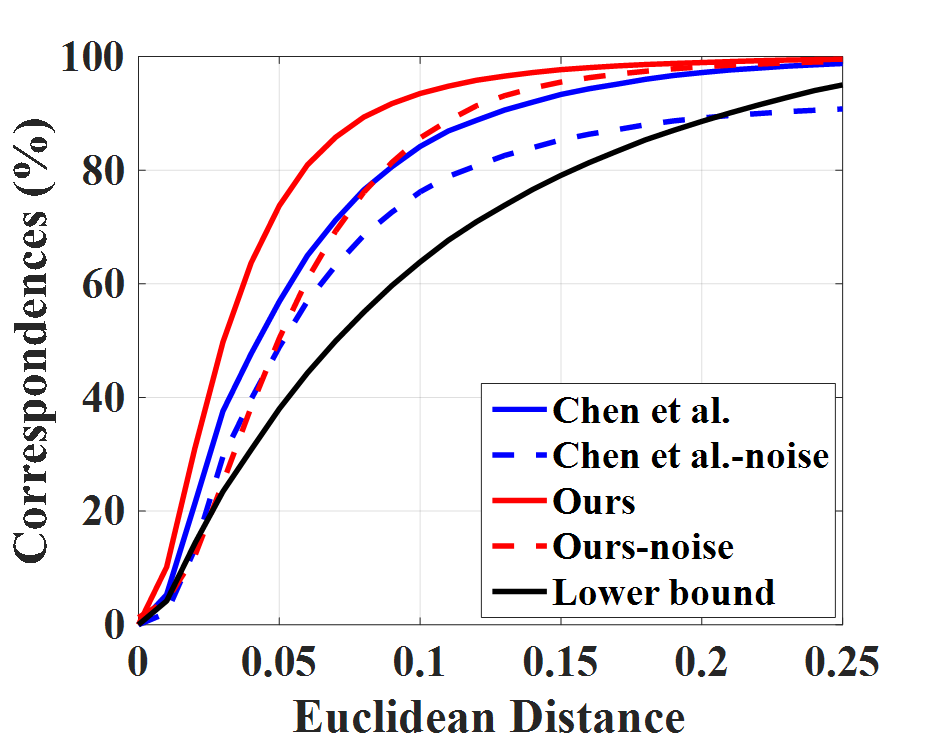}} 
\hspace{-2mm}
\subfigure[]{\label{fig:ablation_study1_a}\includegraphics[trim=2 4 4 2,clip, height=34mm]{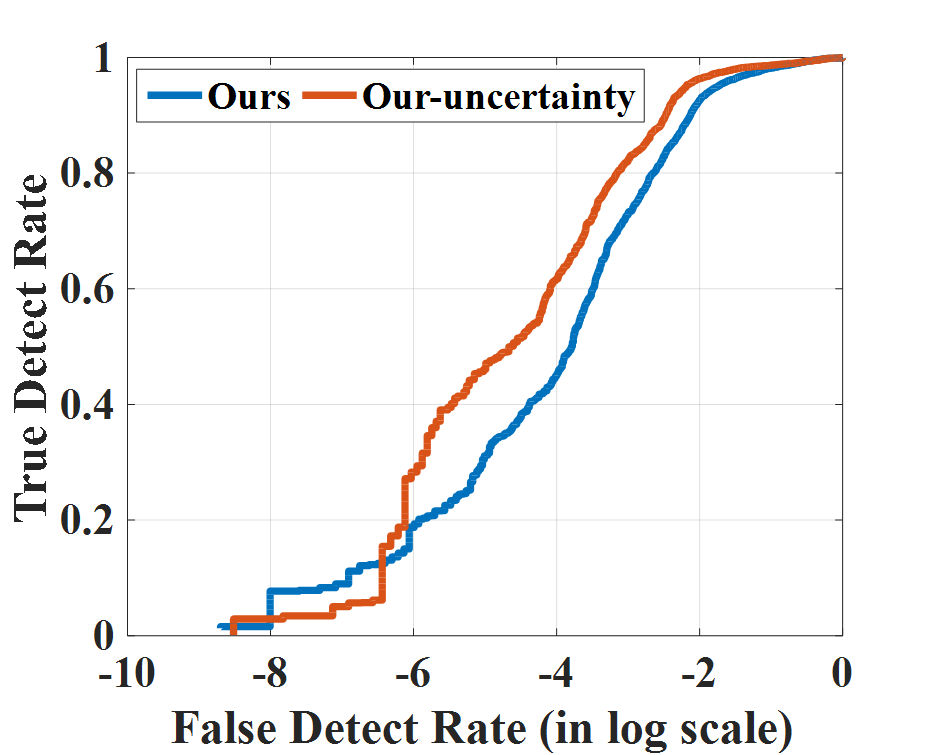}} 
\vspace{-2mm}
\caption{ (a)  Additional semantic correspondence results for the chair category in BHCP. (b) ROC curve of the non-existence of correspondence detection.}
\label{fig:ablation_study1}
\end{figure}

In the aligned setting, one naive approach to semantic correspondence is to find the closest point $q$ on another $3$D shape given a point $p$ in one shape. 
We report the accuracy of this approach as the black curve in Fig.~\ref{fig:ablation_study1_noise}. 
Clearly, our accuracy is much higher than this ``lower bound", indicating our method doesn't rely much on the canonical orientation. 
To further validate noisy real data, we evaluate the Chair category with additive noise $\mathcal{N}(0,0.02^2)$ and compare with Chen~\emph{et~al.}~\cite{chen2020unsupervised}. 
As shown in Fig.~\ref{fig:ablation_study1_noise}, the accuracy is slightly worse than testing on clean data. However, our method still outperforms the baseline on noisy data.

\begin{figure}[t]
\centering    
\includegraphics[trim=0 0 0 0,clip, width=88mm]{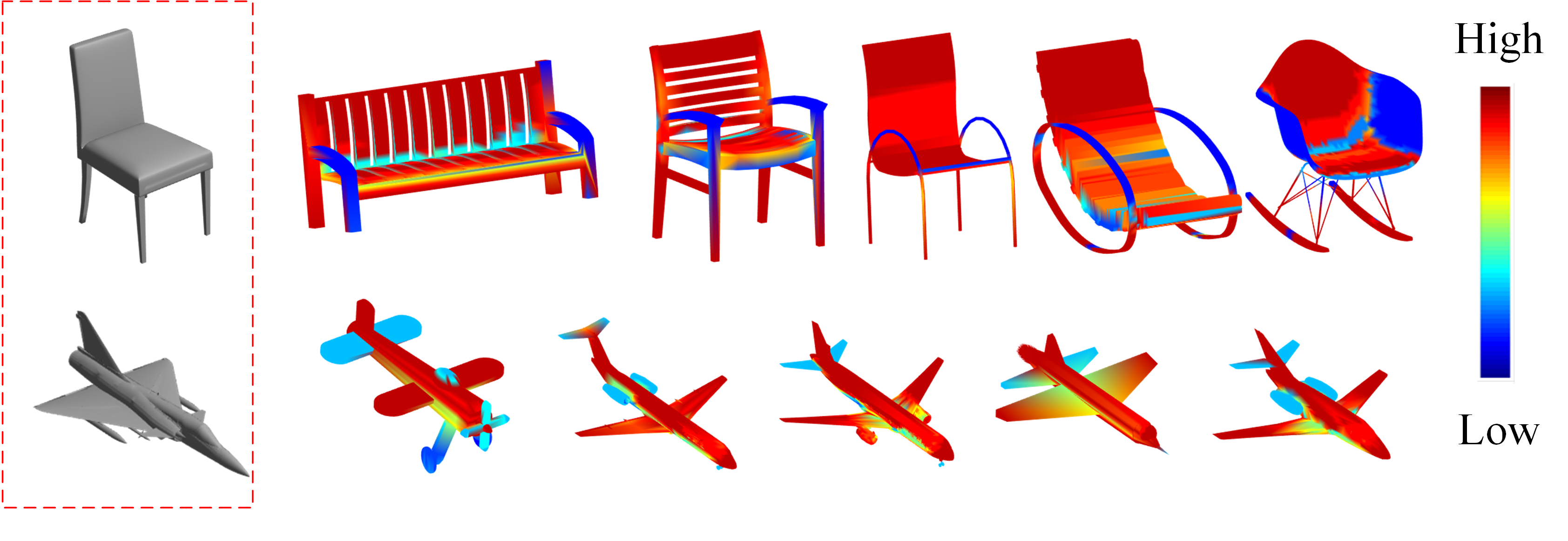} 
\caption{Visualization of the confidence score. The confidence score maps show the probability around corresponded points between the target shape (red box) and its pair-wise source shapes.}
\label{fig:corr_maps}
\end{figure}

%-----
\Paragraph{Visualization of Correspondence Confidence Score}
To further visualize the correspondence confidence score, we provide confidence score maps for some examples. As shown in Fig.~\ref{fig:corr_maps}, the confidence score can show the probability around corresponded points between the target shape (red box) and its pair-wise source shapes. For example, for the source shapes with arms, we can clearly see the confidence scores of the arm part is significantly lower than other parts. 

%-----
\Paragraph{Detecting Non-Existence of Correspondences}
Our method can build dense correspondences for $3$D shapes with different topologies, and automatically declare the non-existence of correspondence. 
The experiment in Fig.~\ref{fig:semantic} cannot fully depict this capability of our algorithm because no semantic point was annotated on a non-matching part. 
Also, there is no benchmark providing the non-existence label between a shape pair. 
 We thus build a dataset with $1,000$ paired shapes from the chair category of ShapeNet part dataset~\cite{yi2016scalable}. 
 Within a pair, one has the arm part while the other does not.
 For the former, we respectively annotate $5$ arm points and $5$ non-arm points based on provided part labels. 
We utilize this data to measure our detection of the non-existence of correspondence. 
Based on our confidence scores, we report the ROCs of both \textbf{Ours} and \textbf{Ours-uncertainty} (AUC: $96.58\%$ vs $97.24\%$) in Fig.~\ref{fig:ablation_study1_a}.
The results show our strong capability in detecting unreliable correspondence.

% -------------Figures of experiment----------------- 

\begin{figure}[t]
\centering    
\includegraphics[trim=5mm 0 0 5mm,clip, width=88mm]{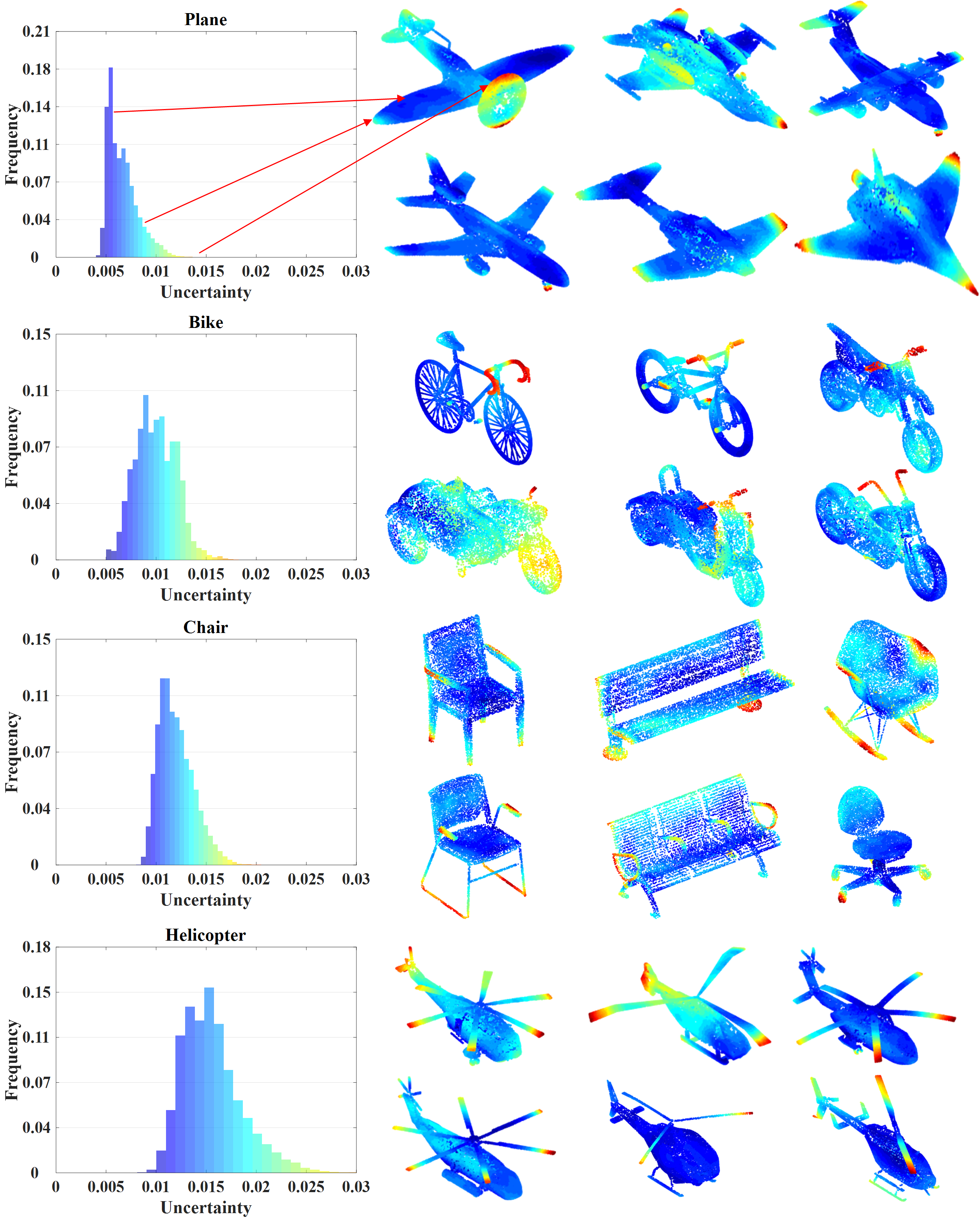} 
\caption{Distribution of point-wise uncertainty in the semantic embedding on the BHCP $4$ categories. Here, ``uncertainty'' refers to the mean value of $\mathbf{o}_{\sigma}^2$ across all feature dimensions. On the right-hand side, we show the uncertainty distribution by shapes. It can be observed that the learned uncertainty increases along with the shape regions with semantic ambiguity, \emph{e.g.}, the arms and legs of chairs, which often differ among instances.}
\label{fig:uncertainty_dis}
\end{figure}

% -------------Figures of experiment-----------------  
 \begin{figure*}[t]
\centering     
\subfigure[]{\label{fig:faust_a}\includegraphics[trim=0 0 0 0,clip, height=38mm]{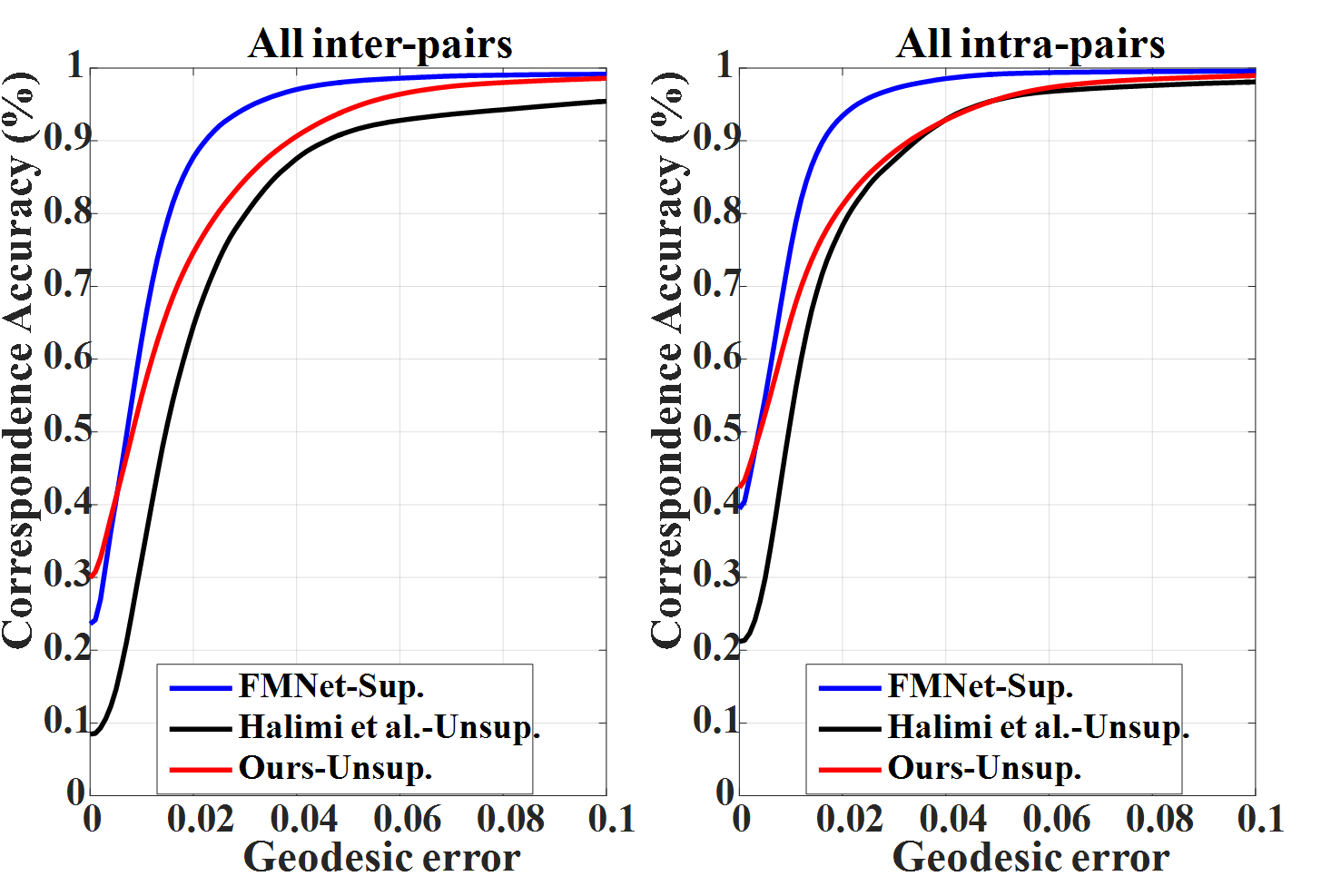}} 
\hspace{1mm}
\subfigure[]{\label{fig:faust_b}\includegraphics[trim=0 0 0 0,clip,height=38mm]{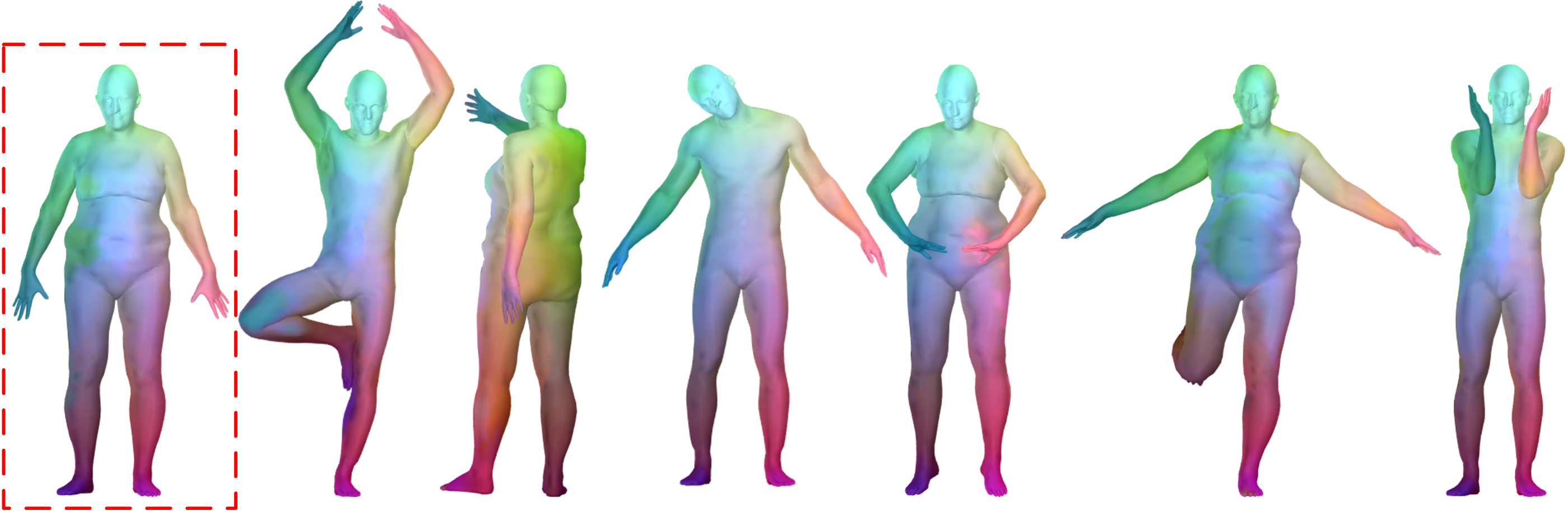}}
\caption{(a) Geodesic error comparison of intra-subject pairs and inter-subject pairs on the FAUST dataset~\cite{bogo2014faust}, for three methods: ours (unsup.), FMNet~\cite{litany2017deep} (sup.) and Halimi~\emph{et al.}~\cite{halimi2019unsupervised} (unsup.) (b) One target shape and $6$ pairwise corresponded source shapes.}
\label{fig:faust}
\end{figure*}

\begin{figure}[t]
\centering    
\includegraphics[trim=0 0 0 0,clip, width=88mm]{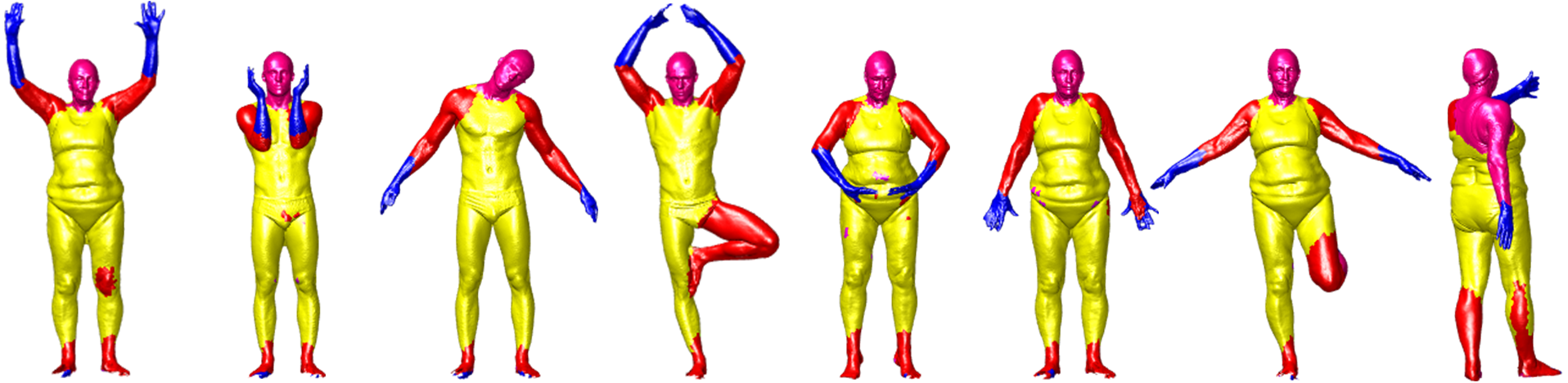} 
\caption{Unsupervised segmentation results of the proposed method on $3$D human shapes.}
\label{fig:faust_seg}
\end{figure}

\begin{figure}[t]
\centering     
\includegraphics[trim=0 0 0 0,clip, width=88mm]{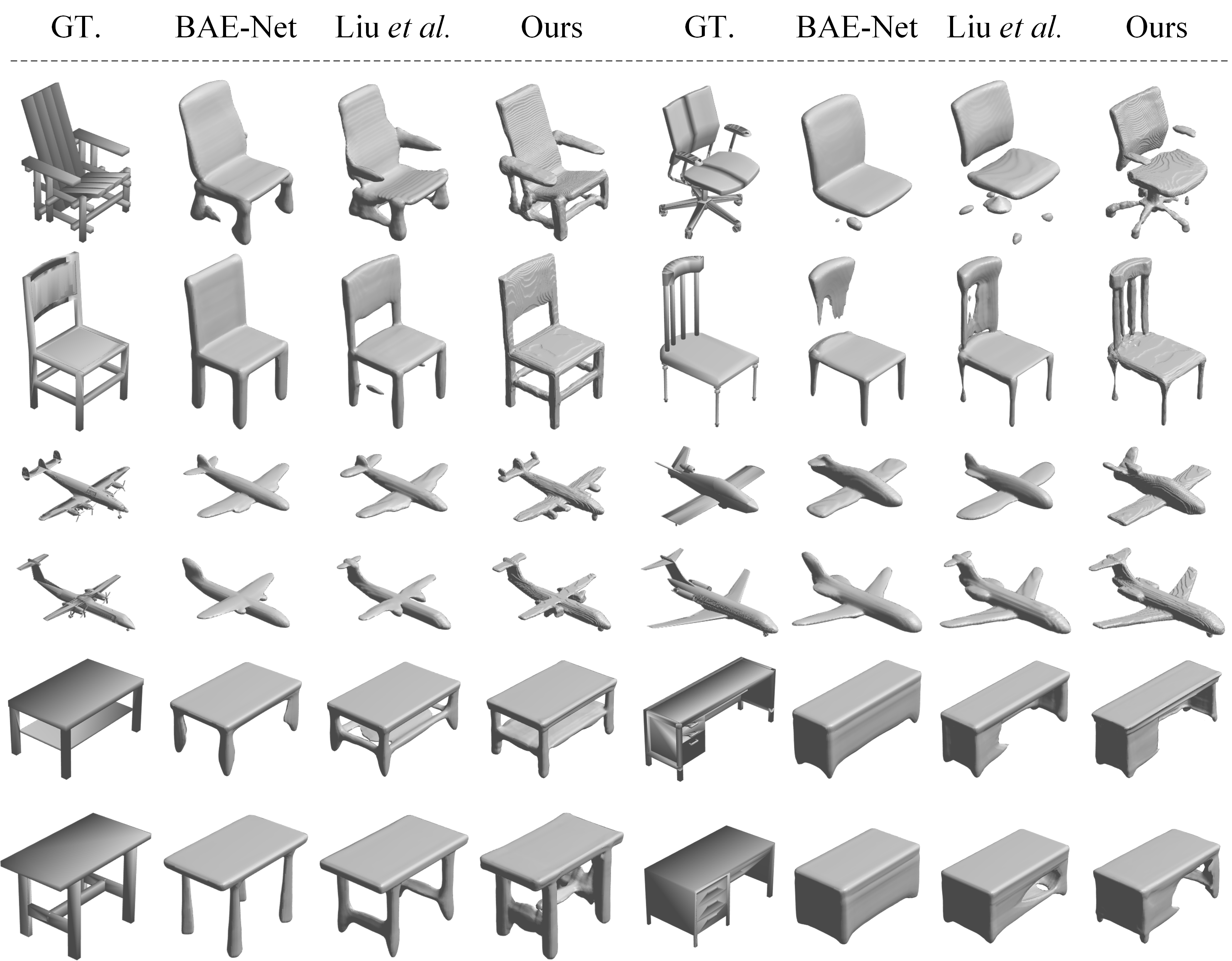} 
\caption{Shape representation power comparison. Our reconstructions closely match the ground-truth (GT.) shapes than BAE-Net~\cite{chen2019bae} and Liu \emph{et al.}~\cite{liu2020correspondence}.}
\label{fig:shape_rep}
\end{figure}
% -------------------------------------------------  

% -------------Figures of experiment-----------------  
\begin{figure*}
    \centering
    \resizebox{1\linewidth}{!}{
    \includegraphics[width=\linewidth]{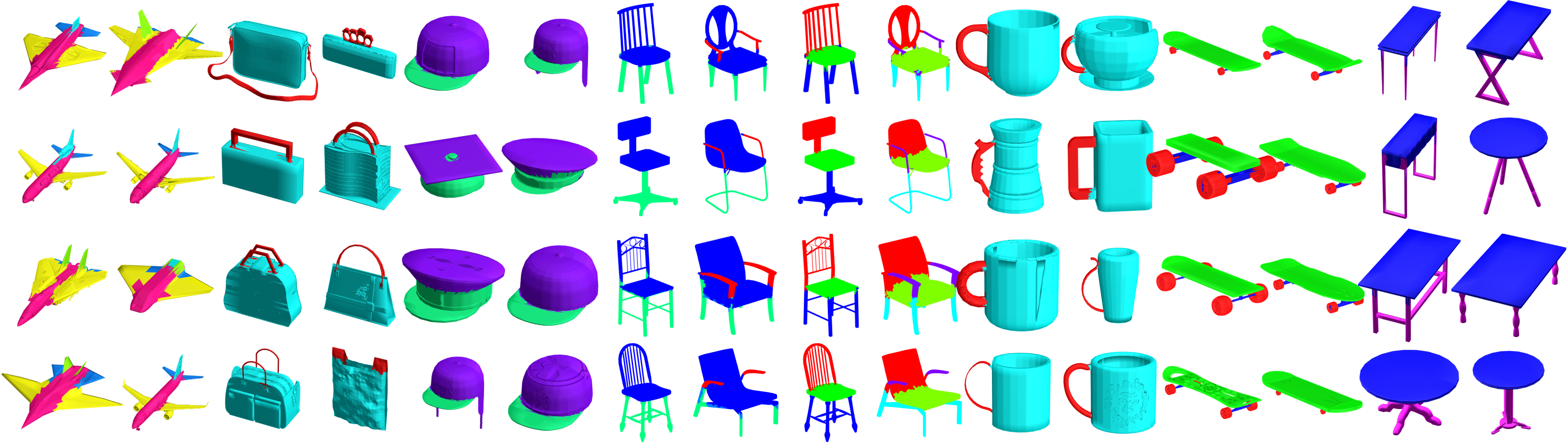}
    }
    \vspace{-2mm}
    \caption{ Qualitative results of our unsupervised segmentation in Tab.~\ref{tab:seg}: $8$ shapes in each of the $8$ categories.}
    \label{fig:seg}
\end{figure*}

\begin{figure*}[t]
\centering     
\subfigure[]{\label{fig:ablation_sTNE_b}\includegraphics[trim=0 12 0 2,clip,height=32mm]{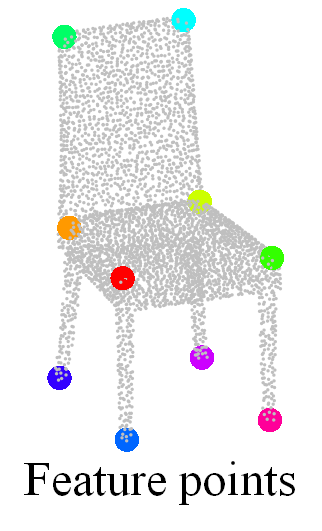}}
\hspace{1mm}
\subfigure[]{\label{fig:ablation_sTNE_c}\includegraphics[trim=0 12 0 0,clip,height=38mm]{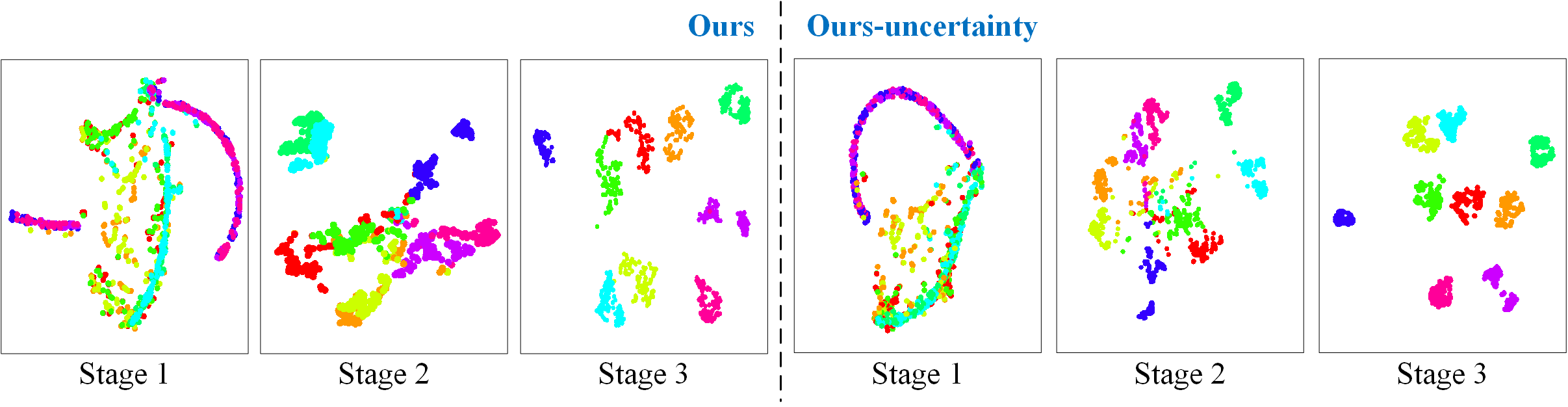}}
\vspace{-2mm}
\caption{(a) $10$ semantic points overlaid with the shape.
(b) The t-SNE comparison of the estimated PEVs over $3$ training stages of models trained with the \textbf{Ours} and \textbf{Ours-uncertainty} settings. Points of the same color are the PEVs of ground-truth corresponding points in $100$ chairs. $10$ colors refer to the $10$ points in (a).}
\label{fig:ablation_sTNE}
\end{figure*}

\begin{figure}[t]
\centering    
\subfigure[]{\label{fig:ablation_study1_e}\includegraphics[trim=2 4 4 0,clip,height=35mm]{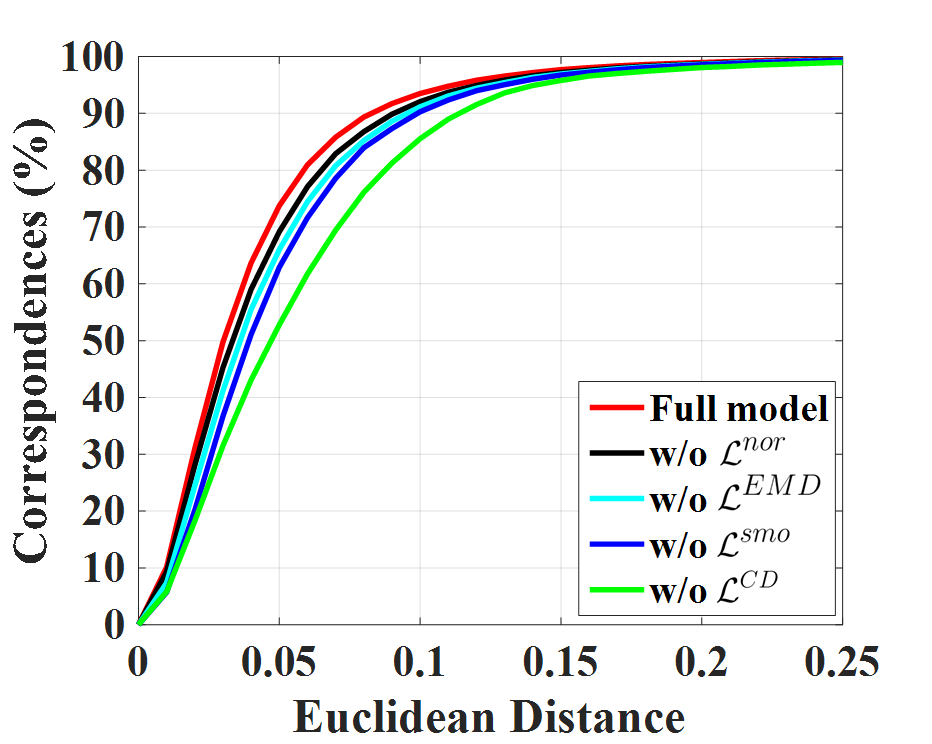}}
\hspace{-2mm}
\subfigure[]{\label{fig:ablation_study1_d}\includegraphics[trim=2 4 4 2,clip,height=35mm]{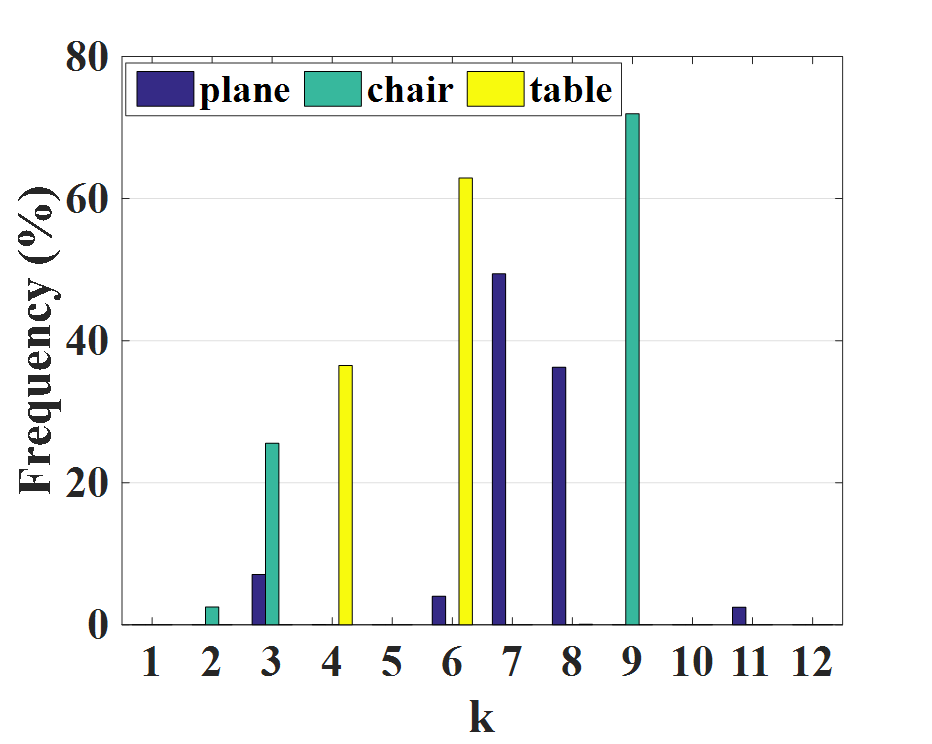}}
\vspace{-2mm}
\caption{(a) $3$D semantic correspondence reflecting the contribution of our loss terms and (b) Active branch distribution of PEVs on three categories (plane, chair, and table). Each branch either represents a shape part or outputs nothing, {\it i.e.}, for the table category, No. $3$ and $10$ branches are active.}
\label{fig:ablation_study3}
\end{figure}

\begin{figure}[t]
\centering     
\subfigure[]{\label{fig:ablation_study1_b}\includegraphics[trim=2 4 4 2,clip,height=35mm]{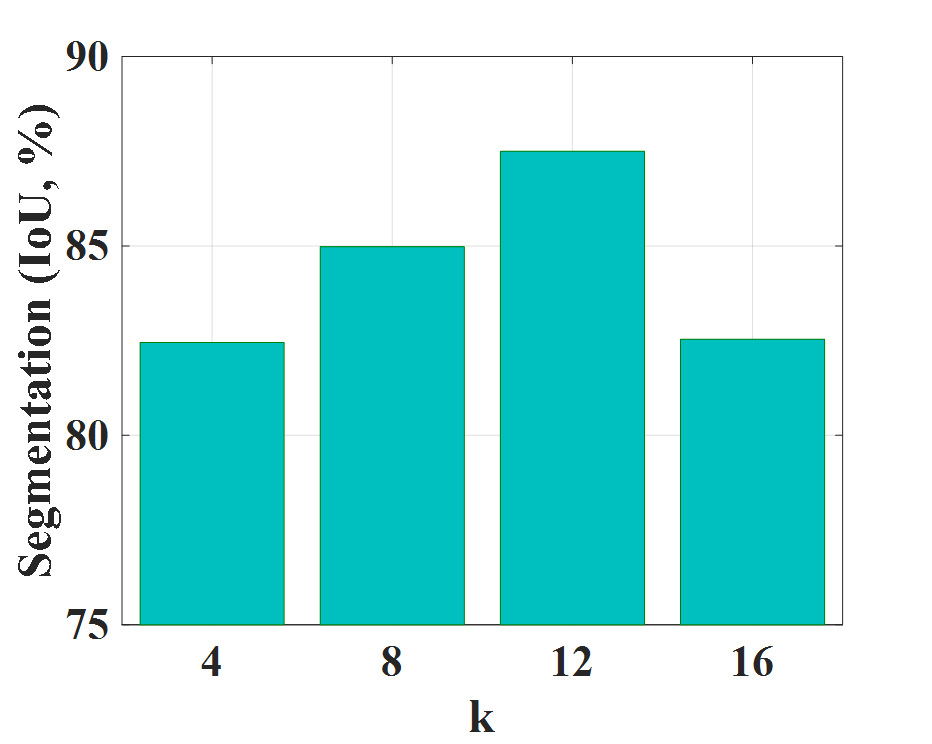}}
\hspace{-2mm}
\subfigure[]{\label{fig:ablation_study1_c}\includegraphics[trim=2 4 4 0,clip,height=35mm]{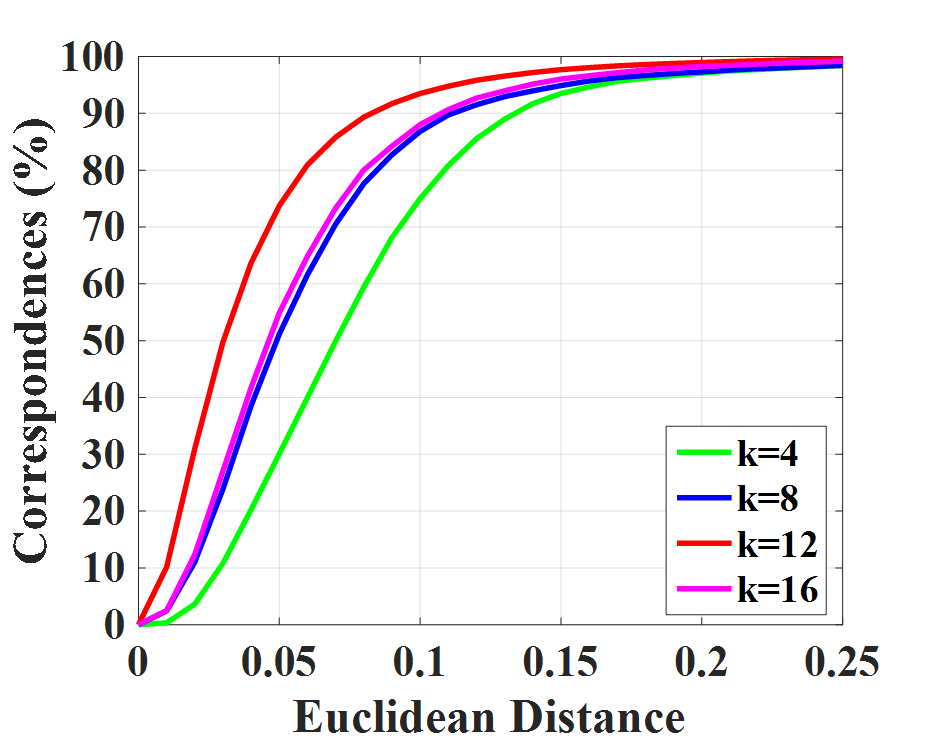}}
\vspace{-2mm}
\caption{(a) Shape segmentation and (b) $3$D semantic correspondence performances on the Chair category over different dimensionalities of PEV ($k$).}
\label{fig:ablation_study2}
\end{figure}

% -------------------------------------------------  
\begin{figure}[t]
\centering    
\includegraphics[trim=-5mm 0 -5mm 0,clip, width=88mm]{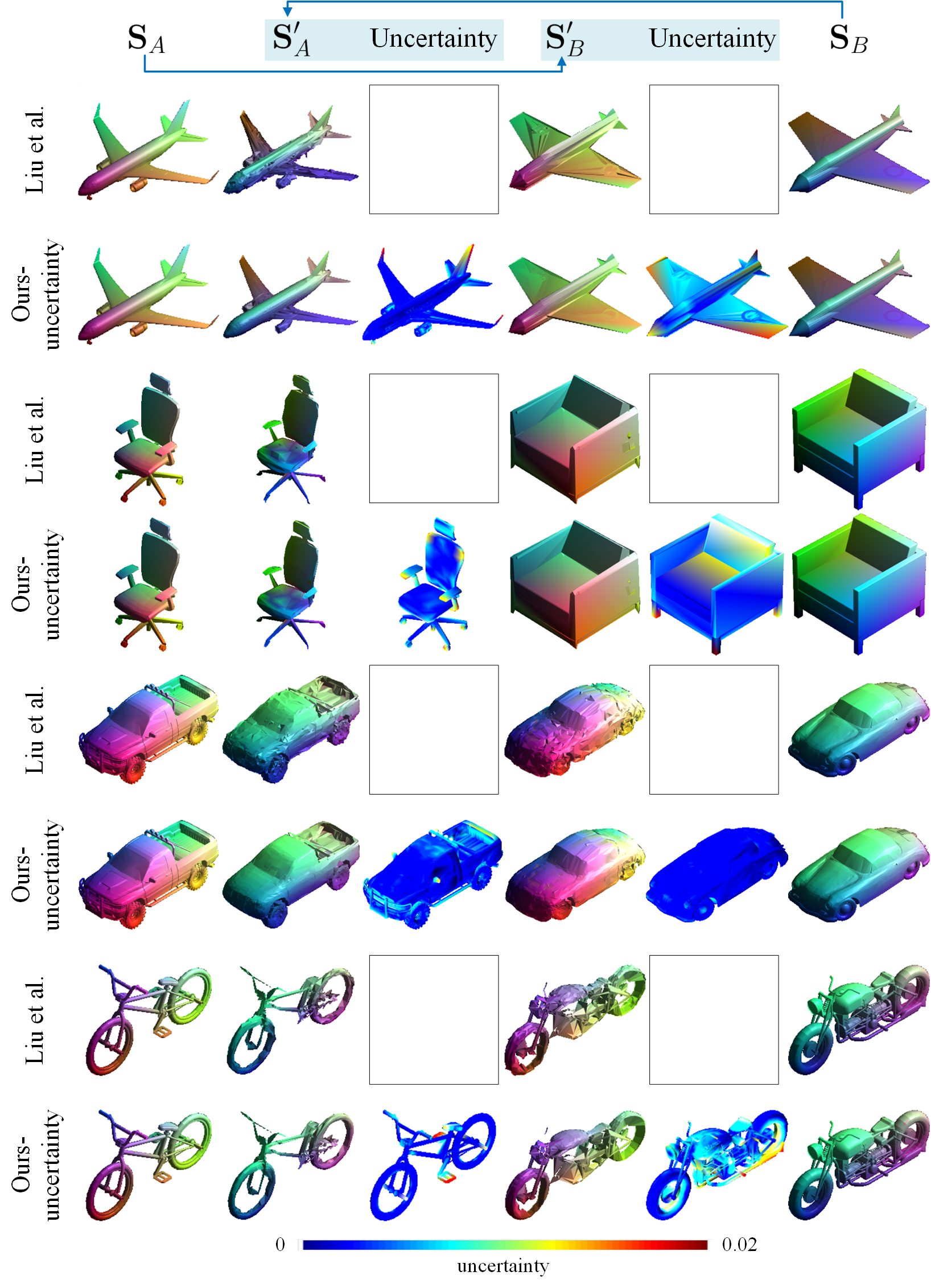} 
\vspace{-2mm}
\caption{Comparison of cross $3$D reconstruction ($\mathbf{S}_{A}\rightarrow \mathbf{S}'_{B}$, $\mathbf{S}_{B}\rightarrow \mathbf{S}'_{A}$) between Liu~\emph{et al.}~\cite{liu2020correspondence} and our method. Corresponding points are assigned the same color in ($\mathbf{S}_{A}, \mathbf{S}'_{B}$) and ($\mathbf{S}_{B}, \mathbf{S}'_{A}$). For our method, we also show the estimated uncertainty, which visualizes the learned variances of cross-reconstructed shapes. Best viewed in zoom-in.}
\label{fig:cross_recon}
\end{figure}
% -------------------------------------------------  

%------------------------------------------------------------------
\subsection{Understand Uncertainty Learning}
To better understand the impact of uncertainty on dense $3$D shape correspondence, we provide the distribution of estimated uncertainty on BHCP $4$ categories in Fig.~\ref{fig:uncertainty_dis}. As can be seen, the uncertainty increases in the following order: planes $<$ bikes $<$ chairs $<$ helicopters. The estimated uncertainty is proportional to the complexity of the object's shape topology, which is intuitive and consistent with semantic correspondence results in Fig.~\ref{fig:semantic}. 
In addition, we visualize the point-wise uncertainty in shapes (Fig.~\ref{fig:uncertainty_dis}). The estimated uncertainty clearly discovers the ``hard'' shape regions in the dense correspondence task, such as the chairs' legs and arms, and bikes' handlebars, which often suffer from large variations in geometric topology. 
Therefore, dense correspondence models with uncertainty learning have two benefits. First, the learned uncertainty can be utilized as a measurement of the complexity of objects' geometric topology in a shape collection.    
Second, the learned uncertainty can also be regarded as a ``confidence indicator'' to identify reliable established point-to-point correspondences. 
%

%------------------------------------------------------------------
\subsection{Dense Correspondence on Human Body}

Although our method is designed to handle challenging man-made or topology-varying objects, we choose to conduct additional experiments for organic shapes for two reasons.
One is that datasets of organic shapes such as human bodies do provide annotations on {\it dense} correspondence.
Thus evaluation of dense correspondence will complement well with our sparse semantic correspondence test in Sec.~\ref{sec:exp_correspondence}.
The other is to evaluate the generalization capability of our method to diverse generic object types.

To this end, we evaluate the FAUST humans dataset~\cite{bogo2014faust}, and compare with two representative SOTA baselines: supervised (FMNet~\cite{litany2017deep}) and unsupervised (Halimi~\emph{et al.}~\cite{halimi2019unsupervised}) methods. 
We follow the same dataset split as in~\cite{halimi2019unsupervised} and~\cite{litany2017deep} where the first $80$ shapes of $8$ subjects are used for training, and a validation set of $20$ shapes of $2$ other subjects is utilized for testing.
The ground-truth densely aligned shapes are used for evaluation. 
As shown in Fig.~\ref{fig:faust_a}, our method, as an unsupervised method, outperforms the unsupervised method~\cite{halimi2019unsupervised}.
Fig.~\ref{fig:faust_b} visualizes the estimated correspondences. We also show the unsupervised segmentation results of some testing shapes in Fig.~\ref{fig:faust_seg}. As can be seen, meaningful and consistent segmentation appears across $3$D human shapes.

%------------------------------------------------------------------

\begin{table*}[t]
\renewcommand\arraystretch{1.1}
\caption{Unsupervised segmentation, shape representation comparisons (IoU/CD-$L_1$) on ShapeNet part. We use \#parts in the evaluation and $k$=$12$ for all models.} 
  \newcommand{\tabincell}[2]{\begin{tabular}{@{}#1@{}}#2\end{tabular}}
  \label{segmentation}
  \centering
  	\resizebox{1\linewidth}{!}{
  \begin{tabular}{l| c|c|c|c|c|c|c|c|c}
    \toprule
    Category (\#parts)     & plane ($3$)     & bag ($2$) & cap ($2$) & chair ($3$) & chair* ($4$) & mug ($2$)  & skateboard ($2$) & table ($2$) & \multirow{3}{*}{Average}\\
     \cline{1-9}
     \tabincell{c}{Segmented \\ parts } & \tabincell{c}{body,tail, \\ wing+engine } & \tabincell{c}{body, \\ handle } & \tabincell{c}{panel, \\ peak } & \tabincell{c}{back+seat, \\ leg, arm } & \tabincell{c}{back, seat, \\ leg, arm } & \tabincell{c}{body, \\ handle } & \tabincell{c}{deck, \\ wheel+bar } & \tabincell{c}{top, \\ leg+support } \\
    \hline
    BAE-Net~\cite{chen2019bae}     &  $80.4/0.020$ &  $82.5/0.059$ & $87.3/0.047$ & $86.6/0.031$ & $83.7/-$ & $93.4/0.028$ & $88.1/0.017$ & $87.0/0.025$ & $86.1/0.032$   \\
    \hline
    Liu~\emph{et al.}~\cite{liu2020correspondence}  &  $81.0/0.015$ &  $85.4/0.044$ & $\textbf{87.9}/0.033$ & $88.2/0.016$ & $86.2/-$ & $\textbf{94.7}/0.023$ & $\textbf{91.6}/0.015$ & $88.3/0.021$ & $\textbf{88.0}/0.024$ \\   
\hline
    Ours  &  $\textbf{82.7}/\textbf{0.009}$ &  $\textbf{85.7}/\textbf{0.035}$ & $87.2/\textbf{0.021}$ & $\textbf{88.6}/\textbf{0.013}$ & $\textbf{86.9}/-$ & $91.9/\textbf{0.014}$ & $88.2/\textbf{0.011}$ & $\textbf{88.7}/\textbf{0.016}$ & $87.5/\textbf{0.017}$ \\   
    \bottomrule
  \end{tabular}
  }
  \label{tab:seg}
\end{table*}

%------------------------------------------------------------------
\subsection{Unsupervised Shape Segmentation}\label{sec:segmentation}
In order to produce $3$D shape segmentation, prior template-based~\cite{kim2013learning} or feature point estimation~\cite{chen2020unsupervised} correspondence methods usually need an additional part template to transfer pre-defined segmentation labels to the estimated corresponded points. However, in contrast to these methods, our framework is able to generate co-segmentation results in an unsupervised manner. For a fair comparison of shape segmentation, we only compare with the SoTA unsupervised method, BAE-Net~\cite{chen2019bae}, which is a \emph{solely} optimized method for shape segmentation.

Following the same protocol~\cite{chen2019bae},
we train category-specific models and test on the same $8$ categories of ShapeNet part dataset~\cite{yi2016scalable}: plane ($2,690$), bag ($76$), cap ($76$), chair ($3,758$), mug ($184$), skateboard ($152$), table ($5,271$), and chair* (a joint chair+table set with $9,029$ shapes). 
Intersection over Union (IoU) between prediction and the ground truth is a common metric for segmentation.
Since unsupervised segmentation is not guaranteed to produce the same part counts exactly as the ground truth, \emph{e.g.}, combining the seat and back of a chair as one part, we report a modified IoU~\cite{chen2019bae} measuring against both parts and part combinations in the ground-truth.
As shown in Tab.~\ref{tab:seg}, our model achieves a higher average segmentation accuracy than BAE-Net and on-par results with our preliminary work~\cite{liu2020correspondence}.  
As BAE-Net is similar to the model of~\cite{liu2020correspondence} trained in Stage $1$, these results show that our dense correspondence task helps the PEV to better segment the shapes into parts, thus producing a more semantically meaningful embedding.
Some visual results of segmentation are shown in Fig.~\ref{fig:seg}.

\subsection{Shape Representation Power of Implicit Function}\label{sec:shape_rep}

We hope our novel implicit function $f$ still serves as an effective shape representation while achieving dense correspondence.  
Hence its shape representation power shall be evaluated.
Following the setting of unsupervised shape segmentation in Sec.~\ref{sec:segmentation}, we first pass a ground-truth point set from the test set to $E$ and extract the shape code $\mathbf{z}$. By feeding $\mathbf{z}$ and a grid of points to $f$, we can reconstruct the $3$D shape by Marching Cubes~\cite{lorensen1987marching}. 
We evaluate how well the reconstruction matches with the ground-truth point set. As shown in Tab.~\ref{tab:seg}, the average Chamfer distance (CD-$L_1$) among branched implicit function (BAE-Net)~\cite{chen2019bae}, our preliminary work~\cite{liu2020correspondence}, and the proposed method on the $7$ categories is $0.032$, $0.024$, and $0.017$, respectively. 
Note that our relative improvement over our preliminary work~\cite{liu2020correspondence} is $29\%$.
This substantially lower CD shows that our novel design of semantic embedding and deep branched implicit function actually improves the shape representation power.
It is understandable that the higher shape representation power is a prerequisite to more precise $3$D correspondence, as shown in Fig.~\ref{fig:semantic}. Additionally, Fig.~\ref{fig:shape_rep} shows the visual quality comparisons of the three categories' reconstructions.

%------------------------------------------------------------------
\subsection{Ablations and Analysis}

%----------------------------------------
\subsubsection{Ablations Study} 

%-------
\Paragraph{Loss Terms on Correspondence} 
Since the point occupancy loss and self-reconstruction loss are essential, we only ablate each term in the cross-reconstruction loss for the Chair category. Correspondence results in Fig.~\ref{fig:ablation_study1_e} demonstrate that, while all loss terms contribute to the final performance, 
$\mathcal{L}^{CD}$ and $\mathcal{L}^{smo}$  are the most crucial ones.
$\mathcal{L}^{CD}$ forces $\mathbf{S}'_B$ to resemble $\mathbf{S}_B$.
Without $\mathcal{L}^{smo}$, it is possible that $\mathbf{S}'_B$ may resemble $\mathbf{S}_B$ well, but with erroneous correspondences locally.

%-------
\Paragraph{Part Embedding over Training Stages}
The assumption of learned PEVs being similar for corresponding points motivates our algorithm design.
To validate this assumption, we visualize the PEVs of $10$ semantic points, defined in Fig.~\ref{fig:ablation_sTNE_b}, with their ground-truth corresponding points across $100$ chairs. 
The t-SNE visualizes the $100\times10$ $k$-dim PEVs in a $2$D plot with one color per semantic point, after each training stage.
The model after Stage $1$ training resembles BAE-Net. As shown in Fig.~\ref{fig:ablation_sTNE_c}, the $100$ points corresponding to the same semantic point, \emph{i.e.}, $2$D points of the same color, scatter and overlap with other semantic (colored) points.
With the inverse function and self-reconstruction loss in Stage $2$, the part embedding shows a more promising grouping of colored points.
Finally, the part embedding after Stage $3$ has well-clustered and more discriminative grouping,
which means points corresponding to the same semantic location do have similar PEVs. 
The improvement trend of part embedding across $3$ stages shows the effectiveness of our loss design and training scheme, as well as validates the key assumption that motivated our algorithm. 
In addition, we compare the t-SNE plot between the \textbf{Ours} and \textbf{Ours-uncertainty} in Fig.~\ref{fig:ablation_sTNE_c}. Here, for the \textbf{Ours-uncertainty}, we apply t-SNE with the mean PEVs ($\mathbf{o}_{\mu}$), 
As can be seen, the part embedding of the \textbf{Ours-uncertainty} has highly well-clustered than the \textbf{Ours} (Stage $3$). It demonstrates that uncertainty learning can further improve the intra-class compactness and inter-class separability in semantic embedding.

%-------
\Paragraph{Dimensionality of PEV} 
As mentioned in Sec.~\ref{sec:impl}, in the final output layer, we utilize a max-pooling operator ($\mathcal{MP}$) to select one branch output and form the final occupancy value. 
Here, we denote the selected branch as an ``active'' branch and Fig.~\ref{fig:ablation_study1_d} shows its distribution of PEVs with $k=12$. As can be observed, the active branch distribution across different categories is random and only a small part of branches is active, which implies that shape segmentation might not require a high-dimensional PEV. 
To verify this, we conduct experiments on the dimensionality of PEV. Fig.~\ref{fig:ablation_study1_b} and~\ref{fig:ablation_study1_c} show the shape segmentation and semantic correspondence results over the dimensionality of PEV. 
Our algorithm performs the best in both when $k=12$.

%-------
\Paragraph{One-hot vs.~Continuous Embedding} 
Ideally, our implicit function, adopted from BAE-Net~\cite{chen2019bae}, should output a one-hot vector before $\mathcal{MP}$, which would benefit unsupervised segmentation the most.
In contrast, our PEVs prefer continuous embedding rather than one-hot. 
To better understand PEV, we compute the statistics of Cosine Similarity (CS) between the PEVs and their corresponding one-hot vectors: $0.972\pm 0.020$ (BAE-Net) vs.~$0.966\pm 0.040$ (ours). 
This shows our learned PEVs are {\it approximately} one-hot vectors. 
Compared to BAE-Net, our smaller CS and larger variance 
are likely due to the limited network capability, as well as our desire to learn a {\it continuous} embedding benefiting correspondence. 

%----------------------------------------

%----------------------------------------
\subsubsection{Expressiveness of Inverse Implicit Function}
Given our inverse implicit function, we are able to cross-reconstruct each other between two paired shapes by swapping their part embedding vectors. Further, we can interpolate shapes both in learned semantic embedding space and maintain the point-level correspondence consistently. 

%-------
\Paragraph{Cross-Reconstruction Performance} We first show the cross-reconstruction performances in Fig.~\ref{fig:cross_recon}. From a shape collection, we can randomly select two shapes $\mathbf{S}_A$ and $\mathbf{S}_B$. Their shape codes $\mathbf{z}_{A}$ and $\mathbf{z}_{B}$ can be predicted by the PointNet encoder. With their respectively generated the mean of PEVs $\mathbf{o}_{\mu A}$ and $\mathbf{o}_{\mu B}$, we swap their PEVs, send the concatenated vectors to the inverse function, and obtain $\mathbf{S}'_{A} = g(\mathbf{o}_{\mu B}$, $\mathbf{z}_{A}), \mathbf{S}'_{B} = g(\mathbf{o}_{\mu A}, \mathbf{z}_{B})$. As compared to our preliminary work~\cite{liu2020correspondence} in Fig.~\ref{fig:cross_recon}, our cross reconstructions are more closely resemble each other in detail. Additionally, this work can produce uncertainty, which can reveal the reliability of the cross reconstructions. 
Here, we provide the cross-reconstruction performance of the car category, where the car-specific model is trained on $659$ shapes of the ShapeNet Part database.

%----------------------------Figures of experiment -------------------
\begin{figure}[t]
\centering     %%% not \center
\includegraphics[trim=-5mm 0 -5mm 0,clip, width=85mm]{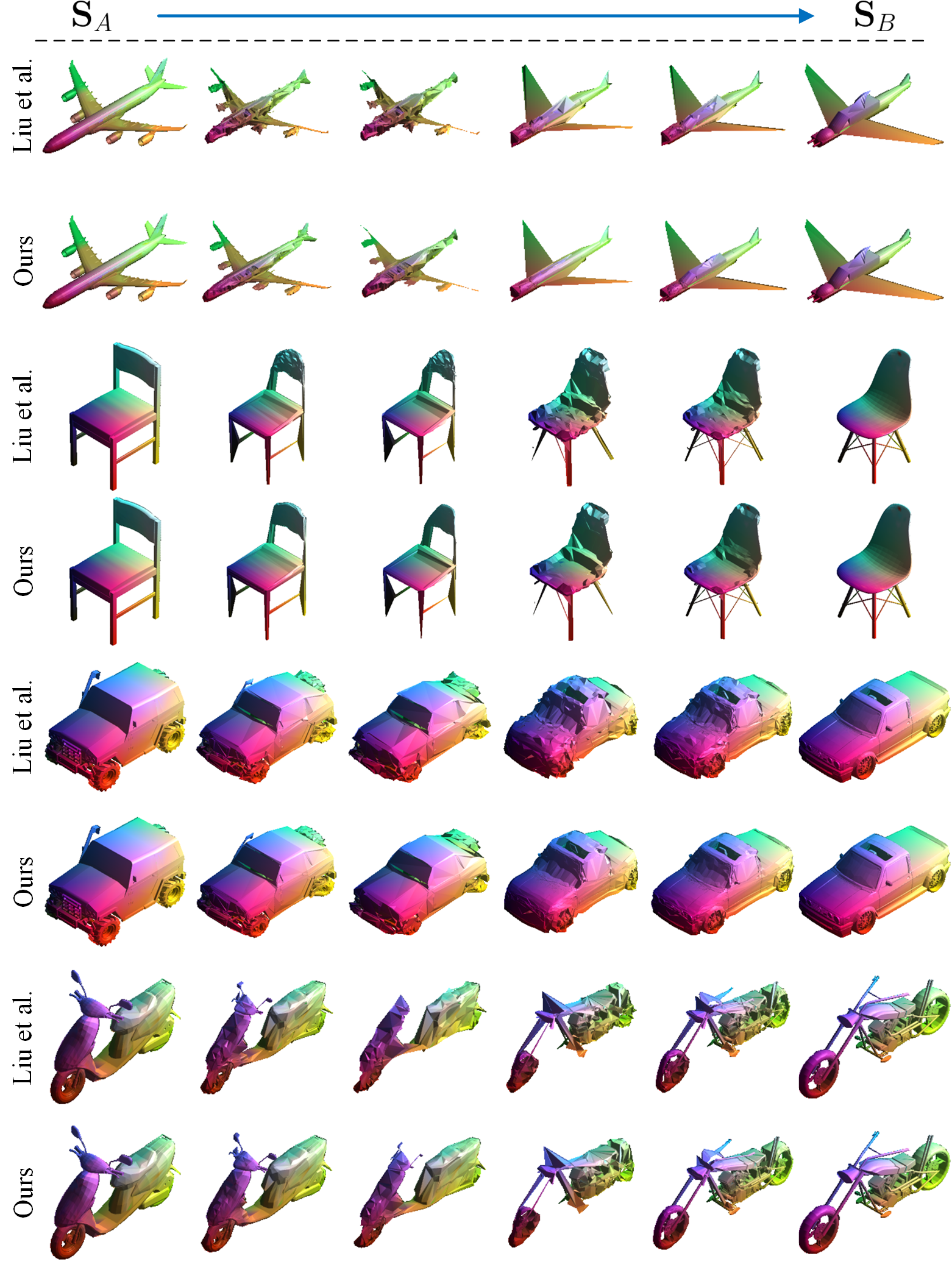} 
\vspace{-2mm}
\caption{Comparison of interpolation in the semantic embedding space between Liu~\emph{et al.}~\cite{liu2020correspondence} and our method. Our interpolations are more smooth and point-to-point consistent than Liu~\emph{et al.}~\cite{liu2020correspondence}. Best viewed in zoom-in.}
\label{fig:inter_latent}
\end{figure}

 \begin{figure}[t]
\centering     %%% not \center
\includegraphics[trim=0 0 0 0,clip, width=85mm]{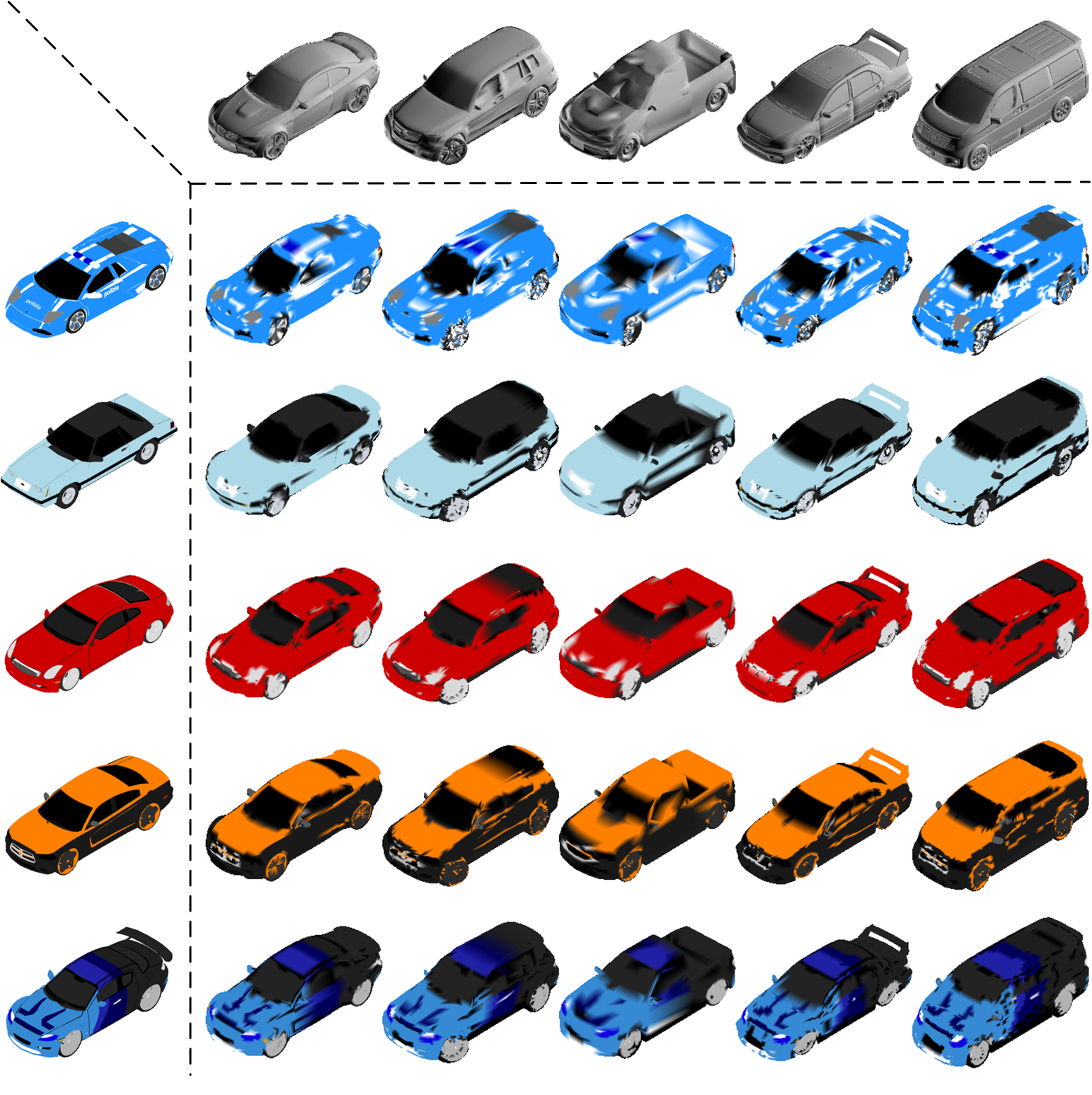} 
\vspace{-2mm}
\caption{Texture transfer from the source shapes ($1^{st}$ column) to the target shapes ($1^{st}$ row)  based on the dense correspondences estimated by the proposed method.}
\label{fig:texture_transfer}
\end{figure}

%-------------------------------------------------------------------

%-------
\Paragraph{Interpolation in the Semantic Embedding Space}
An alternative way to explore the correspondence ability is to evaluate the interpolation capability of the inverse implicit function. 
In this experiment, we interpolate shapes in the latent space. Given two shapes $\mathbf{S}_{A}$ and $\mathbf{S}_{B}$, we first obtain their $\mathbf{z}_{A}$, $\mathbf{z}_{B}$, $\mathbf{o}_{\mu A}$, and $\mathbf{o}_{\mu B}$ by the trained encoder, implicit and inverse implicit functions. The intermediate shape code can be calculated as $\tilde{\mathbf{z}} = \alpha \mathbf{z}_{A} + (1-\alpha) \mathbf{z}_{B}$ ($\alpha \in[0,1]$), and then we send the concatenated vectors ($\tilde{\mathbf{z}}$ and $\mathbf{o}_{\mu A}$) to the inverse function to generate an intermediate cross-reconstructed shape $\mathbf{\tilde{S}}$. Since $\mathbf{S}_{A}$ and $\mathbf{\tilde{S}}$ are point-to-point corresponded, we can easily show the correspondences in the same color. As observed in Fig.~\ref{fig:inter_latent}, our inverse implicit function generalizes well the different shape deformations. Moreover, our interpolations are more smooth and point-to-point consistent than our preliminary work~\cite{liu2020correspondence}. It also demonstrates that the proposed deep branched implicit function and uncertainty learning enhance the discriminative ability of the learned semantic part embedding among different parts of the shape.

%-------
\Paragraph{Texture Transfer}
 As shown in Fig.~\ref{fig:texture_transfer}, based on the correspondences generated by our method, we are able to transfer textures from one shape to another. As can be observed, the texture can be semantically transferred to the correct places in new shapes.
 
%----------------------------------------
\subsubsection{Computation Time}
 Our training on one category ($500$ samples) takes $\sim8$ hours  to
converge with a GTX$1080$Ti GPU, where $1.5$, $2$, and $8$ hours are spent at Stage $1$, $2$, $3$ respectively. 
In inference, the average runtime to pair two shapes ($n{=}8{,}192$) is $0.21$ second including the runtimes of $E$, $f$, $g$ networks (on GPU, $\sim2.1ms$), and neighbor search, confidence calculation (on CPU, $\sim208ms$). The inference time is similar to our preliminary work~\cite{liu2020correspondence} since the deep branched implicit function network only brings an additional $0.6ms$ time cost.

%% file: sec_5_conclusion.tex
\section{Conclusion}\label{sec:conclu}

In this work, we propose a novel framework including an implicit function and its inverse for dense $3$D shape correspondences of topology-varying generic objects. Based on the learned probabilistic semantic part embedding via our implicit function, dense correspondence is established via the inverse function mapping from the part embedding to the corresponding $3$D point. In addition, our algorithm can automatically calculate a confidence score measuring the probability of correspondence, which is desirable for generic objects with large topological variations.  The comprehensive experimental results show the superiority of the proposed method in unsupervised shape correspondence and segmentation.

%% file: sec_6_bio.tex
    \begin{IEEEbiography}[{\includegraphics[width=1in,height=1.25in,clip,keepaspectratio]{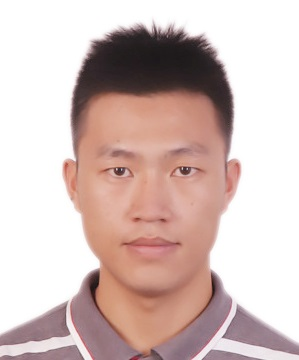}}]{Feng Liu}
        is currently a post-doc researcher in the Computer Vision Lab at Michigan State University. He received the Ph.D. degree in Computer Science from Sichuan University, China in $2018$. His main research interests span the areas of joint analysis of $2$D images and 3D shapes, including $3$D modeling, semantic correspondence, and coherent 3D scene reconstruction. He is a member of the IEEE.
    \end{IEEEbiography}
    
    \begin{IEEEbiography}[{\includegraphics[width=1in,height=1.25in,clip,keepaspectratio]{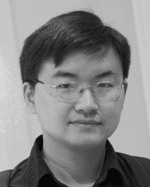}}]{Xiaoming Liu}
        is a Anil K.~and Nandita Jain Endowed Professor and MSU Foundation Professor at the Department of Computer Science and Engineering of Michigan State University. He received the Ph.D. degree in Electrical and Computer Engineering from Carnegie Mellon University in 2004. Before joining MSU in Fall $2012$, he was a research scientist at General Electric (GE) Global Research. His research interests include computer vision, machine learning, and biometrics. As a co-author, he is a recipient of Best Industry Related Paper Award runner-up at ICPR $2014$, Best Student Paper Award at WACV $2012$ and $2014$, Best Poster Award at BMVC $2015$, and Michigan State University College of Engineering Withrow Endowed Distinguished Scholar Award. He has been the Area Chair for numerous conferences, including CVPR, ICCV, ECCV, ICLR, NeurIPS, the Program CO-Chair of WACV'$18$, BTAS'$18$, IJCB'$22$, AVSS'$22$ conferences, and General Co-Chair of FG'$23$ conference. He is an Associate Editor of Pattern Recognition and IEEE Transactions on Image Processing. He has authored more than $150$ scientific publications, and has filed $29$ U.S.~patents. He is a fellow of IEEE and IAPR.
    \end{IEEEbiography}